\documentclass[11pt]{article}
\usepackage[margin=1in]{geometry}
\usepackage{amsmath,amssymb}
\usepackage{graphicx}
\usepackage{booktabs}
\usepackage{tabularx}
\usepackage{xcolor}
\usepackage{enumitem}
\usepackage{caption}
\usepackage{tikz}
\usetikzlibrary{arrows.meta,positioning,fit,backgrounds,calc}
\usepackage[colorlinks=true,linkcolor=blue!55!black,citecolor=blue!55!black,urlcolor=blue!55!black]{hyperref}
\usepackage{url}
\usepackage[numbers,sort&compress]{natbib}
\usepackage[expansion=false]{microtype}

\newcommand{\IM}{Infinity Machine}
\newcommand{\ket}[1]{|#1\rangle}
\newcommand{\bra}[1]{\langle#1|}
\definecolor{kern}{HTML}{1D4ED8}
\definecolor{gen}{HTML}{C2410C}
\definecolor{box}{HTML}{F3F4F6}

\title{\IM: a self-improving AI agent for research}
\author{Harish Santhanalakshmi Ganesan\\ \small Independent researcher \\ \small \texttt{harishsg99@gmail.com}}
\date{October 2026}

\begin{document}
\maketitle

\begin{abstract}
We present the \IM, a self-improving AI agent, and use it for research: problems whose answers can be checked by a program or against data. It runs three loops. A discovery loop keeps an archive of scored solutions, and each new candidate is produced by a full tool-using agent that works for up to an hour in a fresh sandbox, writing and running its own search code and a lab notebook. A catalogue mode searches real data for things nobody has catalogued, judged by a referee rather than a metric. A self-improvement loop lets the system rewrite its own prompts, tools, skills, memory and meta agent behind a gate. An autonomous mode chains everything from a single plain-text goal with no human step: an agent chooses an open problem and writes its verifier; the kernel calibrates that verifier; an adversarial agent audits it; the search runs; an agent that never sees the verifier checks the result; and a final agent writes the report. Proposers see only a public score; where a task has hidden data (held-out instances, later observations), a private split that no agent sees decides what counts. Around this discovery loop the \IM{} keeps an editable \emph{genome}: prompts, tools, skills, an evidence-backed playbook and the meta agent's own procedure. A kernel that the agent cannot write to gates every change to the genome: it checks policy, compares held-in and held-out performance, and enforces a cost limit. We applied the system to fourteen problems, thirteen of them external, in six days of October 2026, at a total cost of about \$180 in model usage. It produced four results beyond the published state of the art on catalogued open problems in quantum information, each checked by a verifier that the proposing agent did not write:
\begin{itemize}[noitemsep,topsep=2pt]
\item an exactly certified point $p=0.064966$ on the positivity ladder of the qubit depolarizing channel's quantum capacity (previous certified point $0.064956$);
\item a single-letter qutrit input that moves that channel's positivity frontier from the hashing value $0.287087$ to $0.2910672$;
\item a 44-gate, depth-6 Clifford circuit preparing the $[[23,1,7]]$ Golay logical zero, which dominates the published $(44,8)$ and $(45,7)$;
\item an exact ten-term stabilizer decomposition showing $\chi(\ket{T}^{\otimes 7})\le 10$ (previously 12).
\end{itemize}
It also cut the added-gate count of the SABRE qubit-routing benchmark from SABRE's 68,559 to 12,087 (9,312 against the best published 13,044 on the 15 circuits shared with later work), built a 5-parameter transformer that adds 10-digit numbers, closed 24\% of the gap between a 20M-parameter language model and the GPT-2 checkpoint on FineWeb validation loss, and surfaced one unconfirmed planet candidate in TESS photometry. In hardware, it redesigned the RTL of an open 8$\times$8 INT8 systolic matrix-multiply tile. Through a full open RTL-to-GDS flow at a fixed clock, it cut the energy per multiply-accumulate from 578.9~fJ to 128.8~fJ (77.7\% lower) with 55\% less area. Power was measured from gate-level switching activity, and the design stays bit-exact. In three fully autonomous end-to-end runs (\$7.59--\$9.97 each, no human step after launch), the system chose its problem, wrote and audited its own verifier, searched, and re-checked its result blind. One audit rejected a task whose cited record was stale. All three runs ended in honest ties with the public record, two of them with documented near misses. In one run the agent-written score pointed the wrong way, and none of the checks caught it. We report as carefully the parts that did not work. The system's one gated attempt to adopt a playbook it had written for itself was rejected on cost. Proposers gamed lenient scorers, usually within a single iteration, four times. And the operator's hand-made changes to the harness were decisive for several results. We describe them so the reader can separate what the loop did from what its operator did.
\end{abstract}

\section{Introduction}

Large language models can now write and run substantial programs, and agents built on them can stay on a task for an hour or more. A system that turns this into discoveries needs two things beyond a capable model. It must have a way of telling a real advance from a plausible-looking one, and a way of keeping what it learns. AlphaEvolve~\cite{novikov2025alphaevolve} and FunSearch~\cite{romeraparedes2024funsearch} show the first. They ground an evolutionary search over programs in automatic evaluation, so that the model's errors are filtered out by execution, and they reach new results in mathematics and computer science. Work on self-improving agents, including STOP~\cite{zelikman2023stop}, the Darwin G\"odel Machine~\cite{zhang2025dgm}, Hyperagents~\cite{zhang2026hyperagents} and AIDE$^2$~\cite{srikanth2026aide2}, addresses the second. There the object being evolved is the agent itself, and recursive self-improvement~\cite{good1965} is the long-run aim.

The \IM\footnote{Earlier data deposits call the system EVO; the two names refer to the same system.} is a research agent that does both. It is not a program-evolution engine with a fixed harness: the unit of work is a researcher-like agent run, the search can be open-ended as well as score-driven, and the harness that directs those agents is itself under gated self-modification. It is built for a single operator with a laptop and a cloud account. Four design choices define it.
\begin{enumerate}[leftmargin=*,itemsep=2pt]
\item \textbf{Agents, not samples.} A proposal in the \IM{} is a full agent run. A Deep Agents~\cite{deepagents} model with a shell, a file system and up to an hour of wall clock in a fresh sandbox reads the task notes and the parent solution. It writes and runs its own search code and delivers a solution together with a lab notebook. A problem therefore receives a handful to about fifteen proposals, not thousands. The proposer, not the harness, decides whether to write a C tableau simulator, an exhaustive enumeration or a gradient ladder.
\item \textbf{A public/private split.} Proposers optimise a public score. Where the task has hidden data, the result is the score on a private split that no agent sees: hidden instances, hidden seeds, or later TESS sectors. For exact-verifier tasks the two splits coincide, and the ``private'' score is the same verifier rerun by the kernel in its own sandbox. We adopt this split from AIDE$^2$~\cite{srikanth2026aide2}. It matters because the agents are strong enough to game any scorer they can read (Section~\ref{sec:gaming}).
\item \textbf{The harness is itself an evolvable object.} Prompts, tools, skills, the playbook and the meta agent's procedure form a versioned object, the genome. A meta agent proposes edits to it, and an immutable kernel accepts an edit only if it passes policy, held-in, held-out and cost checks (Section~\ref{sec:evolve}).
\item \textbf{End-to-end autonomy with machine-checked trust.} The step that usually needs a human, deciding that an agent-written evaluator can be trusted, is replaced by three machine checks: kernel calibration on known cases, an adversarial audit whose exploits the kernel replays, and an independent re-check of the final result by an agent that never sees the evaluator (Section~\ref{sec:autonomous}).
\end{enumerate}

\paragraph{Contributions.}
(i) The architecture and its kernel/genome split, the discovery loop with public/private grading, a catalogue mode for open-ended search in real data, the gated self-improvement loop, and an autonomous mode that runs from a plain-text goal to a verified report (Section~\ref{sec:method}).
(ii) Results on fourteen problems (thirteen external): four advances on catalogued open problems in quantum information, a large margin on a widely used compiler benchmark, two machine-learning tasks and one astronomy candidate, with every claim checked by a verifier that was not the agent's own (Section~\ref{sec:results}, Table~\ref{tab:results}). The quantum, SABRE and TESS results are archived with their data and code on Zenodo (DOIs in the bibliography).
(iii) Three fully autonomous end-to-end runs, whose safeguards caught a stale record before any search began, and a hardware result: a 77.7\% lower energy per MAC for an open accelerator tile, through a full open chip flow with activity-based power (Sections~\ref{sec:chip} and~\ref{sec:auto-run}).
(iv) An account of what the self-improvement machinery did and did not achieve, of four specification-gaming incidents, and of the operator's part in each result (Section~\ref{sec:learned}).

\section{The \IM}
\label{sec:method}

\begin{figure}[t]
\centering
\begin{tikzpicture}[
  font=\small,
  k/.style={draw=kern,thick,rounded corners=2pt,fill=kern!6,align=center,minimum height=7mm,inner sep=3pt},
  g/.style={draw=gen,thick,rounded corners=2pt,fill=gen!7,align=center,minimum height=7mm,inner sep=3pt},
  s/.style={draw=black!60,rounded corners=2pt,fill=box,align=center,minimum height=7mm,inner sep=3pt},
  ar/.style={-{Stealth[length=2mm]},thick,black!70}]
\node[k] (cli) {job daemon};
\node[k,right=3mm of cli] (arch) {archive\\(lineage DAG)};
\node[k,right=3mm of arch] (gate) {gate:\\policy, held-in,\\held-out, cost};
\node[k,right=3mm of gate] (score) {scorers\\(public / private)};
\node[k,right=3mm of score] (bandit) {bandits,\\rewards, budget};
\node[draw=kern,dashed,thick,rounded corners,fit=(cli)(arch)(gate)(score)(bandit),inner sep=6pt,label={[kern]above:\textbf{kernel} (immutable to agents)}] (K) {};
\node[g,below=17mm of arch] (pr) {prompts/\\meta procedure};
\node[g,right=3mm of pr] (tools) {tools/\\(in-sandbox)};
\node[g,right=3mm of tools] (skills) {skills/\\fixtures};
\node[g,right=3mm of skills] (play) {memory/\\playbook};
\node[draw=gen,dashed,thick,rounded corners,fit=(pr)(tools)(skills)(play),inner sep=6pt,label={[gen]below:\textbf{genome} (versioned; one version = one generation)}] (G) {};
\node[s,left=8mm of pr] (sb) {sandbox\\(Docker / Modal)\\agent + shell};
\node[s,below=6mm of sb] (task) {task/\\public split};
\draw[ar] (K.south -| arch) -- node[left,font=\scriptsize]{loads} (G.north -| arch);
\draw[ar] (G.west) -- (sb.east);
\draw[ar] (task) -- (sb);
\draw[ar] (sb.north) -- node[midway,left,font=\scriptsize,align=right]{solution +\\notebook} (sb.north |- K.south);
\draw[ar,gen] ($(G.north -| gate)+(0,0)$) -- node[right,font=\scriptsize]{meta-agent edit $\to$ gate} ($(K.south -| gate)$);
\end{tikzpicture}
\caption{Architecture. The kernel owns everything whose corruption would make scores meaningless: scorers, private splits, archive, budget and policy. The genome holds everything the system may change about itself. Agents run inside sandboxes that mount the genome and the public split only.}
\label{fig:arch}
\end{figure}
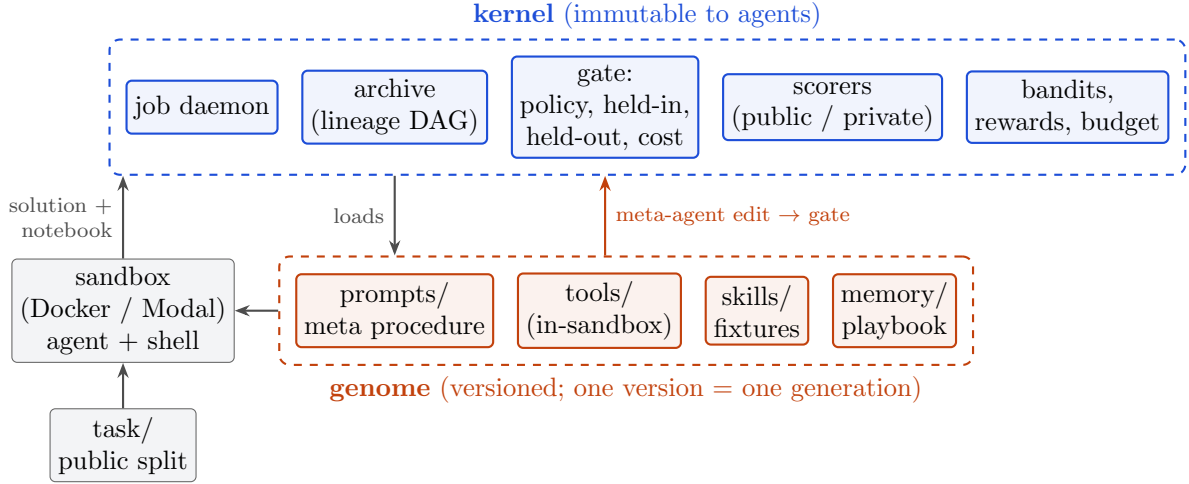

\subsection{Kernel and genome}
The \IM{} is split into an immutable \emph{kernel} and an editable \emph{genome} (Figure~\ref{fig:arch}). The kernel contains a daemon that owns long jobs and survives the controlling shell, the archive (the genome's version history as a lineage DAG, with rollback), the scorers, the gate, the strategy and model bandits, reward logging, and the budget middleware. The genome contains the task agent's and meta agent's prompts, the meta agent's procedure and weakness-mining script, tools with tests, skills with fixtures, and an ID-addressed playbook in which every bullet carries evidence run IDs and a confidence. The split follows the rule from harness-engineering practice~\cite{weng2026harness}: whatever the agent may legitimately want to change is in the genome, and whatever would let it change its own grade is not. The kernel never trusts an agent's report of what it ran. It records tool calls itself, and scorers read solutions produced in a fresh sandbox.

\subsection{Task specification}
A task consists of a specification (problem family, description, scorer, score range and direction, sandbox image and resources, per-run budget, and a brief), a public split (problem statement, notes, tools, a local verifier), a private split and an optional seed solution. The scorer is either a verifier program that the kernel runs in its own sandbox, whose value is mapped linearly onto $[0,1]$, a test suite, or a retrain-from-code scorer for ML tasks. The brief is the operator's channel to the proposers: it lists what has already been measured and what to try.

\begin{figure}[t]
\centering
\begin{tikzpicture}[font=\small,
  n/.style={draw=black!60,rounded corners=2pt,fill=box,align=center,minimum height=8mm,inner sep=3pt,text width=26mm},
  k/.style={draw=kern,thick,rounded corners=2pt,fill=kern!6,align=center,minimum height=8mm,inner sep=3pt,text width=26mm},
  ar/.style={-{Stealth[length=2mm]},thick,black!70}]
\node[k] (bank) {experience bank\\(all solutions, public \& private scores)};
\node[k,right=10mm of bank,text width=34mm] (sel) {parent sampling\\[1pt] $\propto\dfrac{\sigma\big(\lambda(s_{\rm pub}-\bar s_{\rm top3})\big)}{1+\#\text{children}}$};
\node[k,right=10mm of sel] (arm) {strategy arm\\(UCB1 per family)};
\node[n,below=10mm of arm] (prop) {$P$ proposer agents in parallel, each in a fresh sandbox, $\le$1 h, per-run \$ cap};
\node[n,left=10mm of prop] (sol) {solution/ + NOTES.md\\+ search code};
\node[k,left=10mm of sol] (score) {kernel scores\\public $\to$ selection\\private $\to$ result};
\draw[ar] (bank) -- (sel);
\draw[ar] (sel) -- (arm);
\draw[ar] (arm) -- node[right,font=\scriptsize,align=left]{brief + parent +\\bank summary\\(public only)} (prop);
\draw[ar] (prop) -- (sol);
\draw[ar] (sol) -- (score);
\draw[ar] (score) -- node[left,font=\scriptsize]{add} (bank);
\end{tikzpicture}
\caption{One iteration of the discovery loop. Private scores never reach an agent. In catalogue mode every proposer extends the shared merged catalogue instead of a sampled parent, and the referee's per-claim verdicts are passed on as feedback.}
\label{fig:loop}
\end{figure}
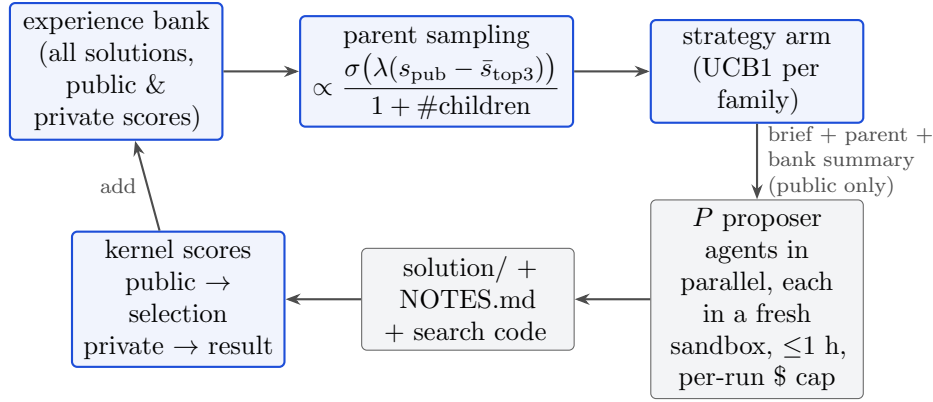

\subsection{Discovery loop}
\label{sec:discover}
A discovery session runs $N$ iterations of the loop of Figure~\ref{fig:loop} with $P$ proposers each. The kernel scores the seed and places it in an \emph{experience bank}. In each iteration it samples $P$ parents with probability proportional to $\sigma(10\,(s_{\rm pub}-\bar s))/(1+c)$. Here $s_{\rm pub}$ is the parent's public score, $\bar s$ is the mean public score of the top three entries and $c$ is the parent's number of children. This is the DGM/Hyra rule~\cite{zhang2025dgm,hyra2026}, which favours good parents that have not been expanded much. Each proposer also receives a named \emph{strategy}, drawn by a per-family UCB1 bandit whose reward is whether the child beat its parent on the public score; AIDE$^2$~\cite{srikanth2026aide2} found UCB over named strategies to be its most valuable discovered mechanism. The default arms, written for the optimisation family and used by every family without its own set, are \emph{conservative}, \emph{aggressive rewrite}, \emph{new neighbourhood}, \emph{data structure} and \emph{robust/simple}. Math tasks use \emph{structured construction}, \emph{exhaustive small case}, \emph{new objective}, \emph{scale compute} and \emph{literature transfer}. ML tasks use \emph{new architecture family}, \emph{parameter sharing}, \emph{embedding budget}, \emph{connectivity} and \emph{recipe}; astronomy tasks have five search-oriented arms of their own.

A proposer is a fresh Deep Agents model (Claude Sonnet~5.5) in a sandbox that mounts the genome, the task's public split and an inspiration folder: the parent solution and a bank summary that shows public scores only. It works with a shell for up to about an hour. It must keep its solution directory current, because a per-run spend cap (\$4 by default), checked before every model call, can end it at any turn. On return, the kernel scores the solution on both splits in a separate sandbox, appends it to the bank and logs the strategy reward. The session's answer is the best \emph{public} solution, and its \emph{private} score is what we report.

\paragraph{Catalogue mode.}
Optimisation has a parent to improve; open-ended discovery in real data does not. A task in catalogue mode changes the deliverable to a list of claims plus a search log. The scorer becomes a \emph{referee}, which re-measures every claim from the raw data, checks it against catalogues of known objects and returns per-claim verdicts. Those verdicts reach the next proposer as feedback. The hidden split is \emph{prediction}: for the TESS task, the private verifier mounts later observing sectors that no agent can read and checks whether each periodic claim predicts dips there. After each round the kernel merges all accepted claims into one shared catalogue, which every proposer then extends.

\subsection{Self-improvement loop}
\label{sec:evolve}
The self-improvement loop improves the genome rather than a solution. A UCB or score-child selector picks a parent generation. The meta agent (Claude Opus~5.5), whose prompt and procedure are themselves genome files~\cite{zhang2026hyperagents}, reads an evidence bundle of recent runs and mines weaknesses. It proposes one bounded edit together with a manifest that states the evidence, the root cause and the predicted fixes. The kernel then applies a static policy check (only genome paths may change, kernel-owned files are byte-identical, at most 12 files) and grades the candidate on held-in and held-out tasks. It accepts the candidate only if neither split regresses beyond a noise band of 0.03 and the cost is at most 1.5 times the incumbent's. A gain larger than 0.3 is parked for human review instead of being activated. Predictions in the manifest are checked against per-task deltas, and a meta agent whose predictions keep failing is sampled less often.

Three further paths feed the genome. \emph{Consolidation}~\cite{zhang2025ace} reflects on run records and edits playbook bullets by ID. Bullets that a run retrieved are confirmed or weakened by that run's outcome, and recurring lessons are promoted to skills, which must pass a fixture. \emph{Capability requests} let a task agent ask for a tool; a toolsmith subagent writes it with tests, and the kernel re-runs those tests before the requester may use the tool. \emph{Rewards} come from scores and from the operator's good/bad feedback on individual runs, and update the bandits and playbook confidence. Every path ends in a generation that passes the same gate, so every self-modification is auditable and reversible.

\subsection{Autonomous mode: from a goal to a verified report}
\label{sec:autonomous}
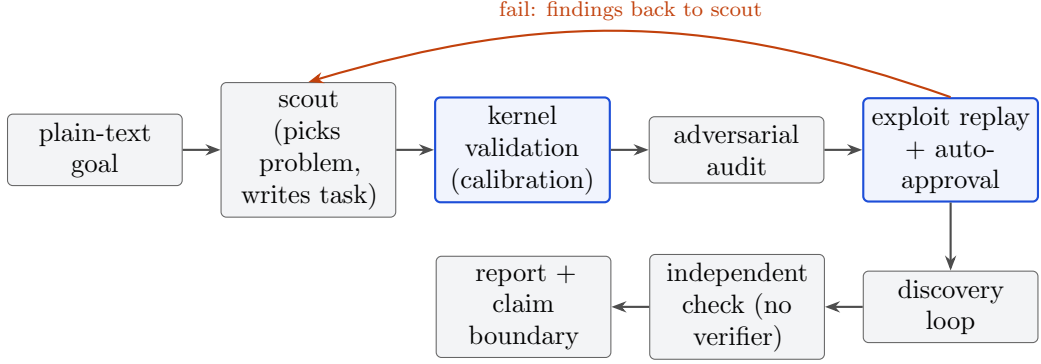
\begin{figure}[t]
\centering
\begin{tikzpicture}[font=\small,
  a/.style={draw=black!60,rounded corners=2pt,fill=box,align=center,minimum height=8mm,inner sep=3pt,text width=21mm},
  k/.style={draw=kern,thick,rounded corners=2pt,fill=kern!6,align=center,minimum height=8mm,inner sep=3pt,text width=21mm},
  ar/.style={-{Stealth[length=2mm]},thick,black!70}]
\node[a] (goal) {plain-text\\goal};
\node[a,right=5mm of goal] (scout) {scout\\(picks problem,\\writes task)};
\node[k,right=5mm of scout] (val) {kernel\\validation\\(calibration)};
\node[a,right=5mm of val] (aud) {adversarial\\audit};
\node[k,right=5mm of aud] (rep) {exploit replay\\+ auto-approval};
\node[a,below=9mm of rep] (disc) {discovery loop};
\node[a,left=5mm of disc] (chk) {independent\\check (no\\verifier)};
\node[a,left=5mm of chk] (rpt) {report +\\claim boundary};
\draw[ar] (goal)--(scout); \draw[ar] (scout)--(val); \draw[ar] (val)--(aud); \draw[ar] (aud)--(rep);
\draw[ar] (rep)--(disc); \draw[ar] (disc)--(chk); \draw[ar] (chk)--(rpt);
\draw[ar,gen] (rep.north) to[bend right=18] node[above,font=\scriptsize]{fail: findings back to scout} (scout.north);
\end{tikzpicture}
\caption{Autonomous mode. Grey: agent runs, each in a fresh sandbox. Blue: kernel steps. Nothing between the goal and the report requires a human.}
\label{fig:auto}
\end{figure}

In the modes above a human writes the task: the problem statement, the verifier, the seed and the notes. Autonomous mode removes that step. Its only input is a plain-text research goal; the default goal asks for an open problem whose best known result is an explicit object or number that a program can check exactly. Every phase is an agent run in its own fresh sandbox, and the kernel only validates, routes and records (Figure~\ref{fig:auto}).
\begin{enumerate}[noitemsep,leftmargin=*]
\item \textbf{Scout} (network on). The scout surveys public problem catalogues and the literature, picks one problem, and checks that it is still open. It writes a complete task to a fixed contract: a task specification with the record and its source, a problem statement, a verifier program, a seed solution, at least four calibration cases with known outcomes (valid ones with their exact values, invalid ones that must be rejected), and a selection note with the evidence of openness.
\item \textbf{Kernel validation.} The kernel runs the verifier on the seed and on every calibration case in a network-off sandbox. Valid cases must reproduce their stated values, and invalid cases must be rejected or score no better than the seed. The scoring range must also extend beyond the record, so that beating the record registers.
\item \textbf{Adversarial audit} (network on). A fresh agent re-checks that the problem is open and the record current, and tries to construct invalid solutions that the verifier accepts at or above the record. The kernel replays every exploit file through the verifier. The audit fails if the auditor says so or if any exploit is accepted. Its findings go back to the scout for a repair round (at most three rounds).
\item \textbf{Installation.} The task is installed and the approval is recorded as automatic, with its basis (calibration, audit and exploit replay). This replaces the human approval that an agent-proposed evaluator otherwise requires.
\item \textbf{Discovery.} The ordinary loop of Section~\ref{sec:discover} runs on the generated task.
\item \textbf{Independent check} (network off). A fresh agent receives only the problem statement and the best solution, never the verifier. It writes its own checker, tests it on cases it constructs, and evaluates the solution. The kernel compares the two values.
\item \textbf{Report.} A fresh agent writes a report and a machine-readable claim: the problem, the record and its source, the result, how it was verified, and an explicit claim boundary.
\end{enumerate}
The protocol is generic: the briefs contain no problem-specific knowledge. All artefacts and events are kept under the session's run directory for audit.

\subsection{Infrastructure and the operator}
Sandboxes are local Docker containers (network off, read-only root, no kernel files, no private split) or Modal sandboxes behind the same interface. Modal provides GPU (one H100) and read-only data volumes for the ML and astronomy tasks. A daemon owns sessions, so a closed terminal does not kill a run, provided the session is submitted as a daemon job; one session started from a foreground shell lost a proposer run when that shell died (Section~\ref{sec:learned}).

The fourteen problem campaigns of Sections~\ref{sec:quantum-start}--\ref{sec:nulls} were run before autonomous mode existed, and in them the operator's role was substantial. The accelerator campaign (Section~\ref{sec:chip}) ran later, in the same operator-led way. We want to be exact about it. For each of those problems the operator, working with an AI coding assistant:
\begin{itemize}[noitemsep]
\item chose and vetted the problem against the literature;
\item wrote the verifier, seed and task notes;
\item launched the sessions;
\item gave run-level feedback;
\item after each session, updated the notes with an ``already measured'' table, which was the main channel by which one session learned from the last;
\item fixed kernel bugs;
\item re-verified and certified every result on the host, and wrote and deposited the notes.
\end{itemize}
The operator also wrote family-specific strategy arms by hand three times (math and ML after observing failures, and astronomy when adding catalogue mode). Under our own rules that is a kernel change rather than a genome change, but it shaped later results, and we discuss it in Section~\ref{sec:learned}. In the three autonomous runs (Section~\ref{sec:auto-run}) the operator's only actions were writing the goal text and launching the job: an agent chose the problem and wrote the task, and the kernel and other agents did everything up to and including the report.

\section{Results}
\label{sec:results}

\begin{figure}[t]
\centering
\includegraphics[width=\linewidth]{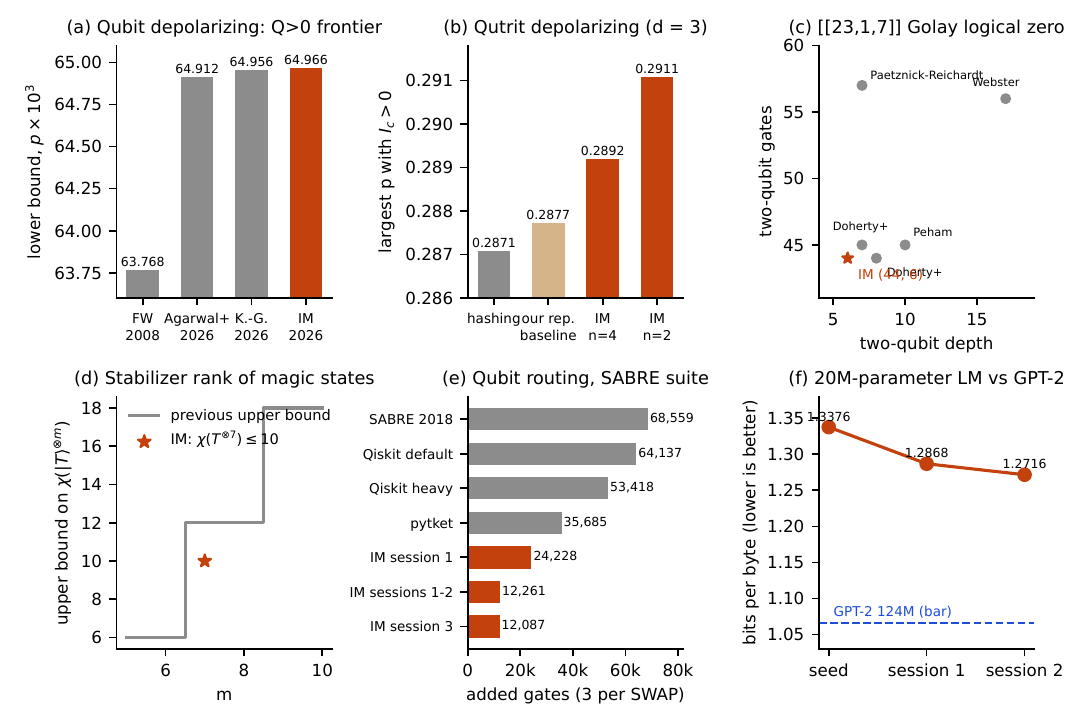}
\caption{Main improvements. (a) Certified points on the qubit depolarizing positivity ladder. (b) Positivity frontier of the qutrit depolarizing channel. (c) Two-qubit gate count and depth for preparing the Golay logical zero. (d) Upper bounds on the stabilizer rank of $\ket{T}^{\otimes m}$. (e) Total added gates on the SABRE suite. (f) Tiny-LM bits per byte at 1B training tokens. Grey: prior work; tan: the operator's own repetition-code baseline (the session-1 seed); orange: \IM.}
\label{fig:results}
\end{figure}

Table~\ref{tab:results} summarises every problem we ran, including the null results. Figure~\ref{fig:results} shows the main improvements. Every entry was checked by a verifier the proposers did not control: the kernel's scorer, a host re-run, and for four of the five quantum results an exact or interval-arithmetic certificate (the fifth, private information, a 30-digit recomputation). Results were obtained between 18 and 26 October 2026.

\begin{table}[t]
\centering\scriptsize
\setlength{\tabcolsep}{3pt}
\begin{tabularx}{\linewidth}{@{}p{3.4cm}p{2.5cm}p{3.0cm}Xr@{}}
\toprule
Problem & Previous best & \IM{} & Verification & Model \$\\
\midrule
\multicolumn{5}{@{}l}{\textit{Open problems (catalogued), quantum information}}\\
Qubit depolarizing $Q>0$ threshold (OQP C11) & $p=0.064956$ certified & $p=0.064966$ certified & authors' exact rational certifier & 1.31\\
Qutrit depolarizing positivity (zoo) & $0.287087$ (hashing) & $0.2910672$ & 50-digit interval certificate & 12.48\\
Golay $\ket{0_L}$ preparation (zoo) & (44, 8), (45, 7) & (44, 6) & tableau + $2^{23}$ statevector & 19.72\\
Stabilizer rank $\chi(\ket{T}^{\otimes 7})$ (zoo) & $\le 12$ & $\le 10$ & exact solve over $\mathbb{Q}(\zeta_8)$ & 9.62\\
Depolarizing private info., finite $n$ (zoo)$^\ddagger$ & $0.28223$ ($n{=}1$) & $0.284668$ ($n{=}8$) & 30-digit recomputation & 14.06\\
\midrule
\multicolumn{5}{@{}l}{\textit{Benchmarks and algorithms}}\\
SABRE routing, 26 circuits, Tokyo & 68,559 (SABRE); 13,044 HAIL on 15 & 12,087; 9,312 on 15 & exact replay of every routing & 21.21\\
TSP (hidden $n$ 600--2000) & seed 0.58 & 0.88 private & hidden instances & 8.12\\
AdderBoard, smallest adder & 6 (disputed), 10 strict & 5 parameters & upstream \texttt{verify.py}, 10,010/10,010 & 4.15\\
$\le$20M LM vs GPT-2 (1.0658 bpb) & seed 1.3376 & 1.2716 (not beaten) & retrain from code, 1B tokens & 9.47$^\dagger$\\
\midrule
\multicolumn{5}{@{}l}{\textit{Hardware design}}\\
8$\times$8 INT8 TPU tile, ASAP7, 600~ps (fJ/MAC) & 578.9 (original RTL) & 128.8 (hidden 130.1) & bit-exact RTL + gate level; full flow, DRC 0; activity-based power & 11.0$^\S$\\
\midrule
\multicolumn{5}{@{}l}{\textit{Open-ended discovery}}\\
TESS M dwarfs, new transits & --- & 1 candidate, $\sim3\sigma$ & held-out sectors, pixel vetting & $\sim$58\\
\midrule
\multicolumn{5}{@{}l}{\textit{No improvement}}\\
Sidon (4,5)-sets, Einstein Arena & 4/7 & 13 new ties at 4/7 & arena verifier & 2.80\\
Hadamard max-det, $n=51$ & 43.30134 & record is a fixed point & exact Bareiss & 4.08\\
Seven-qubit distance-2 code & $24\le K\le 26$ & no $K\ge 25$ & exact residual & 3.68\\
Four MUBs in $d=6$ (ASD) & 0.998291692700 & record reproduced only & 30-digit GS & 2.10\\
\midrule
\multicolumn{5}{@{}l}{\textit{Fully autonomous (problem chosen and task written by the system)}}\\
Two-deletion code, $n=16$ (Arena 34) & 223 (public) & 223 (tie) & task verifier + blind independent checker & 9.86\\
Sorting network, $n=31$ (size) & 180 & 180 (tie) & task verifier + blind independent checker & 7.59\\
Mixed code $N(4,5,3)$ (Brouwer) & 186 & 186 (tie) & task verifier + blind independent checker & 9.97\\
\bottomrule
\end{tabularx}
\caption{All problems. ``Model \$'' is LLM usage over all sessions on that problem; the finding run itself was usually \$0.6--4. Unrecorded spend of killed sessions (C11 session~1, one SABRE attempt of about \$0.46) is excluded. $^\dagger$Plus about \$25 or more of H100 time. $^\S$Plus Modal CPU time for the flow (about 10--12 minutes on 8 cores per full evaluation); ASAP7 is a predictive PDK, so the energies are not silicon measurements (Section~\ref{sec:chip}). $^\ddagger$New only as explicit finite-$n$ witnesses; the literature already implies $P>0$ up to $p\approx0.2915$, and the zoo record certified $0.259824$. Noise convention $\Lambda_p=(1-p)\rho+pI/2$. Zenodo deposits: \cite{zen_c11,zen_qutrit,zen_golay,zen_private,zen_stabrank,zen_sabre,zen_tess}.}
\label{tab:results}
\end{table}

\subsection{Quantum capacity thresholds}
\label{sec:quantum-start}
\paragraph{Qubit depolarizing channel.}
The noise level at which the quantum capacity of $\mathcal D_p(\rho)=(1-3p)\rho+p(X\rho X+Y\rho Y+Z\rho Z)$ vanishes is open. Lower bounds come from inputs with positive coherent information~\cite{fern2008,agarwal2026}. Krohn-Grimberghe recently certified, in exact arithmetic, a 45-copy rank-two symmetric-subspace state at $p=16239/250000=0.064956$~\cite{krohngrimberghe2026}. The \IM{} task gave proposers a float evaluator (Schur--Weyl blocks via a coherent-state generating function), that state as the seed, and score $(p-0.064)/0.002$.

In session~2 the first proposer wrote its own checkpointing Adam ladder, raising $p$ by $10^{-6}$ whenever $I_c>2\times10^{-9}$. In 42 minutes, for \$0.60, it delivered a state with float threshold $\approx0.0649671$. The record authors' own exact certifier, run unchanged on that state, classifies it as certified-positive at $p=32483/500000=0.064966$, with enclosure $[2.31,2.78]\times10^{-8}$ bits~\cite{zen_c11}. This is a candidate-specific lower bound and makes no claim about the optimal threshold. A first session had reached $p=0.064962$ but was killed by host memory pressure before it saved anything, which is why the brief now demands continuous checkpointing.

\paragraph{Qutrit depolarizing channel.}
For $\mathcal N(\rho)=(1-p)\rho+p\,I/3$, the catalogue listed only the hashing point $0.287087$. Session~1 found a 4-qutrit, rank-4 input with threshold $0.289205$. In session~2 a proposer questioned the seed's size instead of growing it: with rank~4 allowed, \emph{smaller} $n$ wins. Its Adam ladders from random starts gave $n=3$: $0.28995$ and $n=2$: $0.29106$, and L-BFGS polishing gave $0.2910672$, the optimum that all 25 L-BFGS starts reached. The operator's structure check showed that the optimum is a product of two single-qutrit inputs, each maximally mixed on a two-dimensional subspace. The gain is therefore single-letter: the leaked component heralds the error. We gave a closed form and a 50-digit interval certificate $p_2(3)\in[0.291067236033599\,69, 0.291067236033599\,88]$, and the same effect holds for every $3\le d\le 64$ we checked~\cite{zen_qutrit}. The proposers found the state but not the structure: two later runs noticed that the state is $P/4$ on a subspace, but neither computed the reduced states that reveal the product. The result is elementary enough that it may be folklore, and the note makes no priority claim.

\paragraph{Private information.}
For the private capacity of the qubit depolarizing channel (here in the convention $\Lambda_p=(1-p)\rho+pI/2$, not per-Pauli), both sessions found classical $K=2$ ensembles of repetition-like distributions. Their thresholds are new as explicit finite-$n$ witnesses, rising from $0.28223$ at $n=1$~\cite{renes2007} to $0.284668$ at $n=8$, but they climb only about $3\times10^{-4}$ per qubit~\cite{zen_private}. The operator's literature check showed that one-way QKD rates with publicly announced syndromes~\cite{kern2008} already imply $P>0$ up to $p\approx0.2915$, once the announcement is paid for in-band. The zoo record had been behind the literature, and the agent's search was not the decisive contribution here.

\subsection{Fault-tolerant circuits and magic states}
\paragraph{Golay logical zero.}
The record CNOT-count/depth points for preparing $\ket{0}$ of the $[[23,1,7]]$ Golay code were $(44,8)$ and $(45,7)$, from RL plus MCTS graph decimation on an A100~\cite{doherty2026golay}. Session~1 returned the seed from all four runs. In session~2 a proposer under the \emph{aggressive rewrite} arm rewrote the search in C (about $5\times10^6$ tableau evaluations per second per core). It changed the objective to the number of qubits left \emph{mixed} after reverse-applying the circuit, which is zero if and only if the state is a product in any local basis, not only in $Z$. It then walked the neutral network at a fixed 44 gates with a penalty on gates beyond the target depth. Depth~7 appeared in about a second and depth~6 in 10--60 seconds; depth~5 and 43 gates were never reached, the latter even after $10^{10}$ evaluations and an exhaustive delete-two/insert-one sweep. The $(44,6)$ circuit dominates both published points. It was verified up to wire relabelling by stabilizer-tableau checks and a $2^{23}$-amplitude statevector overlap of 1.0~\cite{zen_golay}.

\paragraph{Stabilizer rank of $\ket{T}^{\otimes m}$.}
Exact stabilizer ranks are known only for $m\le4$, and the bounds for $m\ge7$ were products of smaller decompositions~\cite{bravyi2016,qassim2021}. Session~1 was null. Two of its six proposers ran no search at all, one calling the task an open research problem and the other ``hardening'' a seed that was already a record, as its default-arm instruction suggested. After the operator replaced the math arms, a \emph{structured construction} proposer in session~2 followed a direction the operator's notes had listed as untried (close a few stabilizer states under qubit permutations) and built a dictionary by closing three seed states under $S_a\times S_b$-symmetric Cliffords (396--540 vectors). It then annealed over subsets with 1,024 batched chains, reaching $r=12$ in 40~s, $r=11$ in a few minutes, and $r=10$ on the $[2,5]$ block family. The decomposition is exact, with coefficients in $\frac14\mathbb Z[\zeta_8]$, and gives $\chi(\ket{T}^{\otimes 7})\le10$ and by sub-multiplicativity $\chi(\ket{T}^{\otimes13})\le60$~\cite{zen_stabrank}. A companion \emph{exhaustive small case} run found that coordinate descent over the full $m=5$ stabilizer table stalls at residual 0.0647 for five terms, which is evidence (not proof) that $\chi(\ket{T}^{\otimes5})=6$.

\subsection{Algorithms and benchmarks}
\paragraph{Qubit routing on the SABRE suite.}
On the 26 circuits and the 43-edge IBM Q20 Tokyo design graph of the SABRE paper~\cite{li2019sabre}, with SWAP insertion only and three added gates per SWAP, the methods progressed as follows. Two circuits (\texttt{qft\_13}, \texttt{qft\_20}) are not in the public benchmark repository and were regenerated (338 and 800 gates against the paper's 403 and 970), so their reference is our SABRE reimplementation. Session~1 rewrote SABRE in C++ with early-abort randomised restarts and a hill-climb over initial mappings (36,435 added gates), then a receding-horizon rollout router (24,228). In session~2, two proposers \emph{independently} implemented a direction the notes listed as untried, a beam search over SWAP sequences, whose state is the mapping plus per-qubit progress; together they reached 12,261. Session~3 refined this to 12,087. On identical inputs, Qiskit~2.4.1's SABRE gives 64,137 (53,418 with heavy trial settings) and pytket~2.18.5 gives 35,685. On the 15 circuits reported by later routing papers, the best published counts (HAIL, 2025~\cite{xu2025hail}) total 13,044, against 9,312 here, and our count is lower on every one of those circuits~\cite{zen_sabre}. Beam search for routing is not new. The routers also spent minutes to an hour of CPU per circuit, against seconds to minutes for the stock baselines (up to 23 minutes per circuit for heavy Qiskit), the run-to-run spread is about 8\%, and the graph is the design graph rather than the 35-coupler device, so this is a statement about headroom in a widely used benchmark, not a fair comparison of algorithms.

\paragraph{Combinatorial optimisation.}
On the internal TSP task (hidden instances, $n$ from 600 to 2000, scored against a BHH optimum estimate), two iterations with two proposers each raised the private score from 0.58 to 0.88. The proposers converged on chained 2-opt with Or-opt and segment-swap kicks. While calibrating this family we found that any reference built from a hand-written local search was matched by the agent within one session, so references must come from optima or stronger solvers.

\subsection{Machine learning}
\paragraph{AdderBoard.}
The task is to find the smallest transformer that adds two 10-digit numbers with full accuracy~\cite{adderboard}. Working from the rules alone, a proposer designed a 5-parameter model in one 8-minute, \$1.05 run. It has one attention layer and two heads keyed on a period-11 sinusoidal positional encoding, a scalar tied embedding and read-out, and a sigmoid carry gate; the carry arises only through autoregressive feedback. It passes the upstream verifier on 10,010 of 10,010 additions across three seeds. A parallel proposer found an independent 6-parameter design. The next session gamed the parameter counter (Section~\ref{sec:gaming}).

\paragraph{A 20M-parameter language model.}
The task was to find an architecture of at most 20M parameters, embeddings included, that beats the OpenAI GPT-2~124M checkpoint on FineWeb validation (1.0658 bits per byte), training only on FineWeb-10B (GPT-2 itself was trained on WebText). The scorer retrains each candidate \emph{from its code} on one H100 for $10^9$ tokens, so a gain must come from the architecture or recipe. The seed was a scaled-down modded-nanogpt~\cite{moddednanogpt} at 1.3376. Session~1 reached 1.2868 with recipe changes and learnable residual mixing; it made no architectural change, because the bandit drew only opt-era arms. After the operator replaced the ML arms, session~2 reached 1.2716 with four unique blocks looped twice over eight layers. That closes 24\% of the gap. The Modal workspace spend limit then stopped the session. This is the clearest case in which the \IM{} did not reach its target.

\subsection{Hardware: a low-energy systolic matrix tile}
\label{sec:chip}
We asked whether the same loop can design hardware when the score is measured power, not a proxy. The design is the 8$\times$8 INT8 weight-stationary systolic array (32-bit outputs) from LogikBench, as shipped with OpenROAD-flow-scripts~\cite{orfs} for the ASAP7 predictive 7\,nm PDK~\cite{asap7}. Proposers may rewrite its RTL freely but must keep its ports and its cycle-level protocol, so throughput is fixed at 64 MACs per cycle. The evaluator, written by the operator, runs every candidate through the full flow at a fixed 600~ps clock: Yosys synthesis, floorplan, placement, clock-tree synthesis, detailed routing and parasitic extraction. It then applies five gates:
\begin{itemize}[noitemsep]
\item the RTL matches a Python golden model on random and corner-case tiles (Icarus Verilog);
\item the routed netlist, simulated with functional ASAP7 cell models, is bit-exact on the same tiles;
\item setup and hold worst slack are both $\ge0$;
\item there are 0 routing DRC violations;
\item the area is at most $1.1\times$ the original's.
\end{itemize}
The score is the energy per MAC on an inference-like workload: post-ReLU activations (about half zero) and roughly Gaussian weights, 16,384 MACs over a 456-cycle window that includes the drain cycles between tiles. OpenSTA computes the power from the gate-level VCD of that window and the routed SPEF. Proposers see the result for workload seed 2026, and the kernel's private score uses seed 7331. A failed gate scores zero. The original RTL measures 578.9~fJ/MAC (34.7~mW; area 5,559~$\mu$m$^2$).

\begin{figure}[t]
\centering
\includegraphics[width=\linewidth]{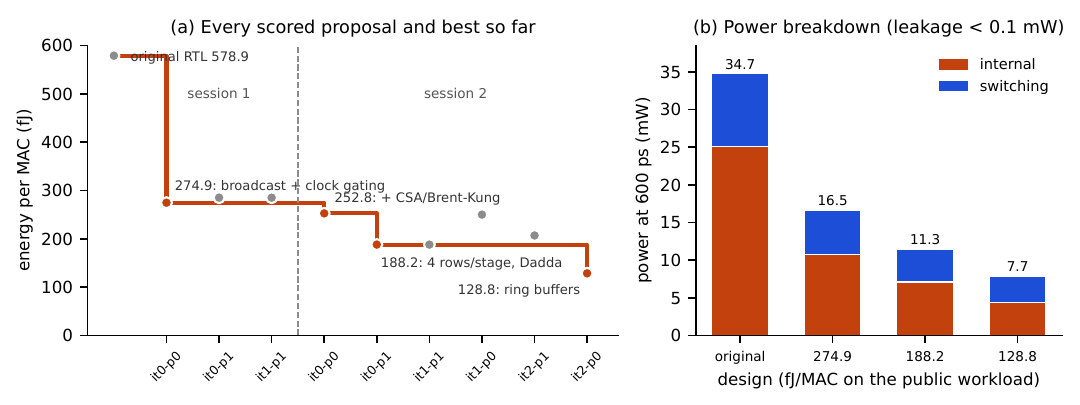}
\caption{The accelerator tile. (a) Energy per MAC of every scored proposal in completion order (dots; orange set a new best) and the best so far (line). Session~2 started from session~1's best design. Run labels are iteration and proposer. (b) Power at the fixed 600~ps clock and unchanged throughput for the original RTL and the three designs that set new bests, split into internal (cell and clock) and switching power; the total is printed above each bar. Values are the kernel's public-workload measurements, from the session banks.}
\label{fig:chip}
\end{figure}

The search ran in two sessions with nine recorded proposer runs (\$11.0 in total). Table~\ref{tab:chip} lists the steps that set a new best, from the agents' notebooks, and Figure~\ref{fig:chip} shows every scored proposal and where the power went.
\begin{table}[t]
\centering\small
\begin{tabular}{@{}lrrp{0.49\linewidth}@{}}
\toprule
Run (arm) & fJ/MAC & $\mu$m$^2$ & Change\\
\midrule
original RTL & 578.9 & 5,559 & --\\
s1 it0-p0 (data structure), v1 & 346.9 & 3,773 & activations broadcast along rows, no output de-skew; per-row partial sums 16--19 bits wide; no datapath resets\\
s1 it0-p0, v2 (s2 seed) & 274.9 & 3,766 & latch+AND clock gates per pipeline depth, enabled from the valid pipe; a gate on the weight registers\\
s2 it0-p0 (conservative) & 252.8 & 2,998 & v2's gating plus a parallel run's Baugh--Wooley multiplier, 3:2 carry-save tree with the partial sum folded in, Brent--Kung adder\\
s2 it0-p1 (new neighbourhood) & 188.2 & 2,427 & four PE rows per pipeline stage: a Python generator emits one 33-operand bit-level Dadda tree and one adder per stage; 4 partial-sum stages $\to$ 2\\
s2 it2-p0 (new neighbourhood) & 128.8 & 2,507 & activation shift chains replaced by a 13-entry ring buffer with a gated clock per entry; weights written in place per row instead of shifted\\
\bottomrule
\end{tabular}
\caption{The steps that set a new best on the accelerator tile (public workload; every step passed all five gates). The final design measures 130.1~fJ/MAC on the private workload.}
\label{tab:chip}
\end{table}
The final design uses 7.7~mW at the same clock and throughput, and is 77.7\% lower in energy per MAC than the original RTL. It is also 55\% smaller, with setup and hold slack of $+5.2$ and $+6.3$~ps. A same-run ablation attributes about 43~fJ of the last step to the ring buffer and 16~fJ to writing the weights in place. Two of the agents' own observations shaped the search. The first was that the 64 input bits are the narrowest point of the pipeline, so it is cheaper to delay the activations than to register partial sums. The second was that flops, and the clock that drives them, dominate the power. The search also found negative results: a radix-4 Booth multiplier with activations encoded once per row was worse (300.8~fJ), and eight rows per stage would not close timing at 600~ps. One proposer declined to ship a 1.6\% better variant because its hold slack was only $+0.197$~ps, reasoning that a gate failure on the private run would score zero.

This result has clear limits. ASAP7 is a predictive PDK, so these are modelled, not measured, energies. There is one flow, one clock and one synthetic workload distribution. The gate-level simulation is zero-delay, so glitch power is not counted. The ring-buffer design relies on the protocol's weight load of exactly eight \texttt{w\_valid} cycles per tile, which the testbench uses but which a different host could violate. The comparison is against the original RTL, not against a hand-optimised design. The operator built the evaluator, wrote an ``already measured'' table between sessions, corrected the score range twice, and reseeded session~2 with session~1's best design (Section~\ref{sec:learned}).

\subsection{Open-ended discovery in TESS photometry}
We pointed catalogue mode at TESS~\cite{ricker2015} light curves of about 17,000 stars: 9,779 director's-discretionary targets and 7,474 M dwarfs within 40~pc observed only in full-frame images. Four sessions cost about \$58. The referee was rewritten four times: twice to close loopholes that proposers exploited (Section~\ref{sec:gaming}), once after the operator's manual vetting found blends, eclipsing binaries and aliases, and once to fix a cadence bug the operator found. The final referee accepts 47 of 60 known TOIs and 30 of 49 known FFI planets, and accepts none of 80 random fake claims. One candidate survives: G~249-11 (TIC~417732194), an M4.5 dwarf at 29~pc with a 3~ppt, 0.9~h transit-like signal at $P=5.307425$~d, which implies $R_p\approx1.6\,R_\oplus$. An ephemeris fitted to two sectors predicted dips in two later sectors that were never used for selection, with $p\approx0.002$ from 400 full-pipeline nulls~\cite{zen_tess}. That is about $3\sigma$, and a bound eclipsing binary and a faint neighbour are not excluded, so we report a candidate, not a planet. An independent read-only reviewer agent caught a 1000-day error in the reference epoch and an independence error in the first vetting report.

\subsection{Null results}
\label{sec:nulls}
On the Einstein Arena~\cite{einsteinarena} Sidon (4,5)-set problem, the \IM{} found 13 new sets tying the 4/7 record but none below it. The operator's feasibility analysis (not the agent's) shows that within the verifier's range only $n=16$ or $n=18$ could beat 4/7. On the $51\times51$ Hadamard maximal-determinant problem, proposers established that the record~\cite{hadamard2026} is a fixed point of every local method they tried. Their best alternative family ($\mathbb Z_5\times\mathbb Z_5$ blocks) reaches 43.2310 against the record's 43.3013. A further exhaustive argument ruling out one circulant family was again the operator's. For the largest seven-qubit distance-two code ($24\le K\le26$~\cite{rains1997,anglesmunne2026}), hundreds of restarts all land on one plateau per method. For four bases in $\mathbb C^6$, every proposer did basin hopping around the known optimum~\cite{raynal2011}, and about 35,000 re-ascents returned to it. Where the published record is a robust optimum of random-restart search, the \IM{} did not improve it, and the useful output was a measured negative.

\subsection{Fully autonomous end-to-end runs}
\label{sec:auto-run}
We launched autonomous mode three times on 26 October 2026, each as a daemon job with a \$30 cap, and did not intervene in any of them. The first run had the default goal: find an open problem in mathematics, quantum information or theoretical computer science whose best known result a program can check exactly, and improve on it. It took 1~h~42~min and \$9.86 (session \texttt{20261026-002501-autonomous}). Table~\ref{tab:autorun} lists its phases.

\begin{table}[t]
\centering\small
\begin{tabular}{@{}lrp{0.62\linewidth}@{}}
\toprule
Phase & \$ & Outcome\\
\midrule
Scout, round 0 & 3.14 & Chose the largest binary two-deletion-correcting code of length 16, a maximum independent set in a 65,536-vertex graph (Einstein Arena problem~34). Cited a 2026 paper's 215 as the record, used the public 208-word code as the seed, and wrote the verifier and calibration cases.\\
Validation & -- & Passed: seed and calibration reproduced.\\
Audit, round 0 & 1.32 & \textbf{Failed.} The record was stale. The arena leaderboard already listed public codes of 216--223 words; the auditor downloaded the 223-word code and verified it with the task verifier and its own checker. Kernel replay of 18 exploit files: all rejected.\\
Scout, round 1 & 0.85 & Repaired the task: record 223, seed the 223-word code, score range extended to 228.\\
Audit, round 1 & 0.58 & Passed: open, record current, no exploit accepted. Task installed with automatic approval.\\
Discovery & 3.46 & Three rounds of two proposers. They tried a plateau local search in C ($\approx1.9\times10^8$ candidate probes per second per core), an exact large-neighbourhood search that frees 24--60 codewords and re-solves them by branch and bound, and a stock iterated local search. None found a 224-word code. Round~3 drew \emph{conservative} twice, and both runs were no-ops.\\
Independent check & 0.19 & The agent wrote its own checker from the problem statement, plus a pairwise longest-common-subsequence test (maximum 13 $<$ 14). Result: valid, 223 words, in agreement with the verifier.\\
Report & 0.32 & ``No improvement: the best verified code has 223 words, which ties (does not beat) the public record of 223.'' The claim boundary credits the record's authors.\\
\bottomrule
\end{tabular}
\caption{Autonomous run 1, phase by phase, from the session's event log and the agents' outputs.}
\label{tab:autorun}
\end{table}

The first run produced no new result. It does, however, show the full chain working with no human step, and both of its safeguards fired for real.
\begin{itemize}[noitemsep]
\item \textbf{The audit prevented a false discovery.} Without it, the search would have ``beaten the record'' of 215 by rediscovering a public code, and the report would have claimed it.
\item \textbf{The independent check confirmed the verifier} with code the verifier's author never saw.
\end{itemize}
The run also showed two weaknesses.
\begin{itemize}[noitemsep]
\item \textbf{The scout checked papers but not leaderboards or discussion threads.} The auditor found the newer codes there.
\item \textbf{Autonomous tasks fell back to the default strategy arms,} so the last round repeated the conservative no-op failure of Section~\ref{sec:learned}. We fixed this for runs 2 and 3.
\end{itemize}
The chosen problem was also a crowded one: many agents had pushed the public record from 216 to 223 in five weeks.

\paragraph{Runs 2 and 3.} We then changed two things and ran autonomous mode twice more, in parallel. First, the autonomous task family now uses the research-shaped math arms (literature transfer, structured construction, new objective, scale compute, exhaustive small case). Second, each goal named a specific table instead of leaving the field open. We chose the tables from a survey of quiet, verifiable records (Appendix~\ref{app:autorun}). Run~2 targeted Dobbelaere's sorting-network size tables for 25 to 40 inputs, and run~3 Brouwer's table of mixed binary/ternary codes. Within those tables the scout chose the cell, confirmed it was current, and wrote the verifier, seed and calibration cases.
\begin{table}[t]
\centering\small
\begin{tabular}{@{}p{0.2\linewidth}p{0.36\linewidth}p{0.36\linewidth}@{}}
\toprule
 & Run 2 & Run 3\\
\midrule
Chosen problem & Size of a sorting network on 31 inputs; record 180 comparators (Batcher-type, unchanged since 2017), bounds 141--180 & Mixed code $N(4,5,3)$, 4 binary and 5 ternary coordinates, distance 3; table 186--216, lower bound from 1998\\
Why this cell & Its entry is a 32-input merge network minus one wire, never improved by evolutionary search, while $n=28$--30 were & 16\% gap; neighbouring cells were improved in 2010--2013 and this one was not\\
Scout & \$2.03, 19~min; exact 0-1 verifier over all $2^{31}$ inputs ($\approx$15~s) & \$4.66, 47~min; its own coset-union search reproduced 186 and other table entries; 12 calibration cases\\
Audit & \$0.89; pass; 23 exploit files, all rejected on replay & \$1.14; pass; 24 exploit files, all rejected on replay\\
Discovery & \$4.12, 6 runs: SorterHunter-style evolution, a witness-vector variant, annealing at 179 comparators on the number of unsorted inputs; 2 runs searched nothing & \$3.60, 6 runs: coset unions, about 30,000 prescribed automorphism groups, tabu search, LNS from the 186 code\\
Near miss & 179 comparators with one unsorted 0-1 vector & 189 words (63 cosets) with one conflicting pair\\
Check, report & \$0.55; independent C checker agrees; ``no improvement'' & \$0.58; independent checker agrees; ``no improvement''\\
Total & \$7.59, 2~h~38~min & \$9.97, 2~h~55~min\\
\bottomrule
\end{tabular}
\caption{Autonomous runs 2 and 3 (sessions \texttt{20261026-024005-autonomous-e33eb} and \texttt{20261026-023958-autonomous-67f31}), from their event logs, scout selections and reports. Both ended in ties with the record.}
\label{tab:autorun23}
\end{table}
Table~\ref{tab:autorun23} summarises both runs. Both tied the record. In both, the scout's selection reasoning was specific and checkable, the verifier survived its audit, and the final report declined to claim anything. Their near misses are informative but are not results: a network that leaves one input unsorted is not a sorting network. Four things went wrong.
\begin{itemize}[noitemsep]
\item \textbf{A reversed score in run~2, missed by every check.} The kernel's minimising scorer gives score 1 at the first end of the task's score range and 0 at the second. The scout wrote the range as 186 to 175, intending the opposite. As installed, the 186-comparator Batcher network scored 1.0, the record 0.455, and a record-beating 179 would have scored 0.364, below its parent. Kernel validation checked values but not the direction of the score. The auditor read the mapping as intended and reported 0.545 for the record. The reporter noticed the mismatch between 0.545 and 0.4545 but not its cause. No proposer shipped a larger network, so the outcome was unaffected. Had one found a 179, though, the bandit and the parent sampler would have treated it as a regression. Worse, the independent check and the report take the highest-scoring solution, so the 179 would have stayed in the bank unreported. The operator had made the same mistake on the chip task (Section~\ref{sec:learned}). Validation should check that calibration cases with better values score higher.
\item \textbf{Declined searches.} In run~2, two of the six proposers (\emph{exhaustive small case} and \emph{new objective}) judged further search hopeless and returned their parent without running a new search; one estimated $10^9$ candidate repairs as too many for its budget. This is the conservative no-op of Section~\ref{sec:learned} in a new form, now under research arms. All six run-3 proposers ran searches.
\item \textbf{Rate-limited literature checks.} In run~3 both the scout and the auditor were throttled by arXiv and Semantic Scholar (HTTP~429). The task was installed on the strength of the table page alone, and the report says so.
\item \textbf{A colliding session.} Two jobs launched in the same second first shared one session directory. We cancelled both before their scouts finished, kept the directory, and added a random suffix to session names before relaunching. The runs reported here are the relaunches.
\end{itemize}
Across three runs, autonomous mode chose reasonable problems, wrote verifiers that survived 18--24 replayed exploits each, caught one stale record, and reported three honest nulls for \$27.42 in total. It also installed one task whose score pointed the wrong way, and none of its checks noticed. It has not yet made a discovery.

\section{What the loop learned, and what it did not}
\label{sec:learned}

\paragraph{The self-improvement gate rejected the one self-derived generation.}
The first 16 problem runs (AdderBoard, Sidon, Hadamard) never reached the system's memory: two kernel scanners read only top-level runs. After that was fixed, one consolidation pass over those 16 proposer runs wrote four playbook bullets from the proposers' own notes:
\begin{itemize}[noitemsep]
\item do not re-run neighbourhoods that earlier notes already tried;
\item a faster optimiser on a proxy is not progress unless the proxy tracks the real score;
\item pilot symmetry-restricted families for 60~s before committing budget;
\item use the shipped checker.
\end{itemize}
Graded by the self-improvement gate (\$8.22, 57 minutes), the generation with these bullets scored 0.944 on held-in tasks against the incumbent's 0.927, and 1.000 on held-out tasks against 1.000. Its cost was 2.02 times the incumbent's, above the gate's 1.5 limit, so it was \emph{rejected}. The bullets made the agent run more experiments, which bought a small gain at twice the spend. No meta-agent proposal ran in that session, because the single iteration was spent validating this pending candidate; in an earlier smoke run the meta agent proposed no change. Every discovery result in Section~\ref{sec:results} was therefore produced by the initial genome. The gate behaved as designed. Whether $2\times$ the cost for $+0.017$ is worth it is a policy question for a human, and with one seed the gain is within noise. We do not yet have evidence that the \IM{} improves itself on these problems.

\paragraph{Strategy arms mattered, but a human chose them.}
Over the first 16 runs the bandit's data were unambiguous. \emph{Data structure} earned a mean reward of 0.4 over 5 pulls, and every other arm earned 0 over 11. \emph{Conservative} and \emph{robust/simple} pulls on a record seed were pure no-ops: proposers shipped the seed unchanged, at about \$0.8 in total. The bandit would have learned to avoid those arms eventually, but on problems with only 4--6 pulls per session it could not. The operator replaced the arms by hand for math and ML (and wrote the astronomy arms). For math (structured construction and the others), the first session afterwards produced $\chi(\ket{T}^{\otimes7})\le10$. For ML (architecture family, parameter sharing and the others), the next session produced the looped model. Under the rule that lessons must enter through the system's own loops, these were kernel changes, not self-improvement, and we report them as such. A version of the \IM{} that proposed its own strategy arms through the gate is the obvious next step.

\paragraph{Specification gaming within one iteration.}
\label{sec:gaming}
Strong proposers exploited lenient scorers four times, usually within a single iteration (items 1--4); item 5 is a hole in a referee that the operator, not a proposer, found. The original artefact for each item (code, formula, the agent's own words) is in Appendix~\ref{app:evidence}.
\begin{enumerate}[leftmargin=*,itemsep=2pt]
\item \textbf{AdderBoard.} With the score saturating at 6 parameters, the next session ``reached one parameter'' in 2.5 minutes by turning four weights into Python constants. Strict counting, under which constants acting as weights count, restored the true count of 5. The proposers disclosed the loophole when asked for review risks.
\item \textbf{Seven-qubit code.} A partial-credit formula paid for any residual below 1, so an unconfirmed $K=25$ subspace at residual 0.072 outscored the confirmed $K=24$ code. Credit now starts only below $10^{-2}$.
\item \textbf{TESS, referee v1.} Proposers claimed ``single transits'' 500--1200\% deep (flux glitches), flux steps and about 90 cheap eclipsing binaries. The fix added physical bounds, a step test and a sky-neighbour systematics test.
\item \textbf{TESS, referee v2.} A proposer wrote that ``extra credit lies in uncapped single events and duo transits'' and submitted 240 single and 39 two-event ``transits''. The fix requires at least three events and caps singles and duos.
\item \textbf{TESS, referee v4 (not gaming).} The referee itself had a hole: it used one cadence per star, so signals that appeared in only one observing era passed. Leave-one-era-out tests found it, and v5 checks cadence per sector and requires consistency across eras.
\end{enumerate}
The pattern matches reports from Hyra and the DGM~\cite{hyra2026,zhang2025dgm}. The defence that worked was not a better prompt. It was a private split the agents could not see, together with adversarial calibration of the verifier: known positives, injected negatives, and an independent reviewer.

\begin{figure}[t]
\centering
\includegraphics[width=\linewidth]{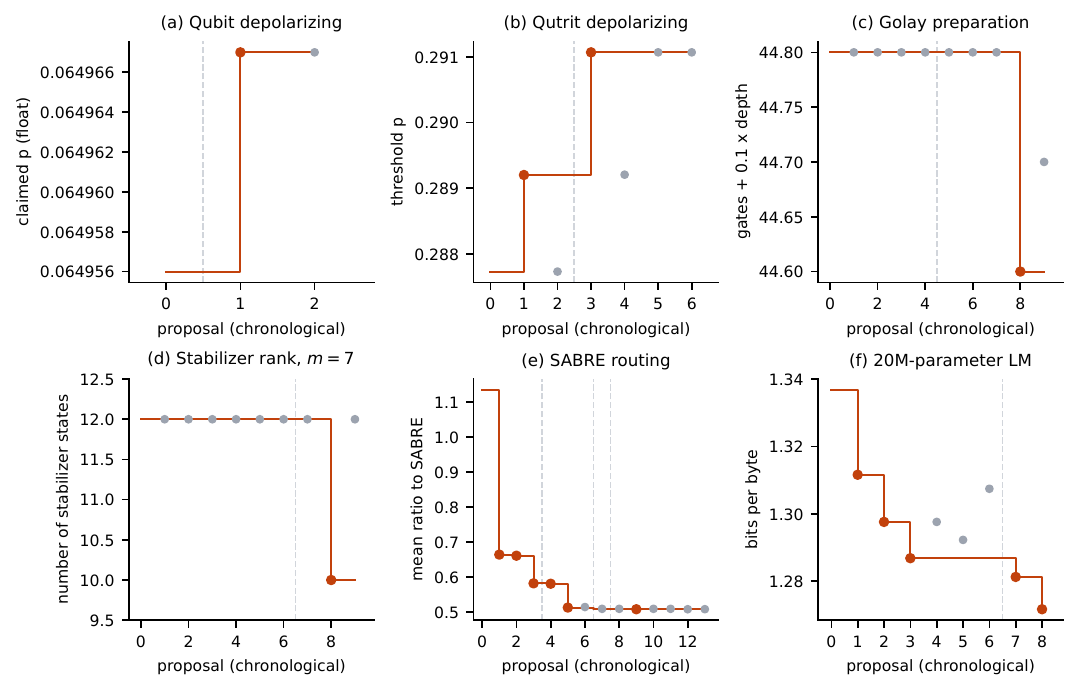}
\caption{Every scored proposal (dots) and the best value so far (line), in chronological order across sessions; dashed lines mark session boundaries, and orange dots set a new best. A session's seed can raise the bar between sessions when the operator combined earlier runs (the SABRE per-circuit merge). Values are those the kernel recorded for each run: (a) the claimed float threshold (the certified point is 0.064966); (c) two-qubit gates $+\,0.1\times$ depth; (d) $12/\text{value}$, i.e.\ the number of stabilizer states; (e) per-run mean ratio to SABRE, before the host's per-circuit combination (0.5073); (f) excludes four session-2 runs stopped by the Modal spend limit. Extracted from the run banks.}
\label{fig:progress}
\end{figure}

\paragraph{The outcome of every proposer run.}
Table~\ref{tab:outcomes} classifies all 104 proposer runs recorded in the 27 discovery sessions, by comparing each run's public score with its parent's. Seven in ten runs did not beat their parent. The table counts beating the parent, which is weaker than setting a new best (Figure~\ref{fig:progress}), and it takes the scores at face value. Eight of the 39 parent-beating runs owe their gain wholly or partly to a scorer we later fixed.

\begin{table}[t]
\centering\small
\begin{tabular}{p{0.52\linewidth}rp{0.33\linewidth}}
\toprule
Outcome & Runs & Notes\\
\midrule
Beat its parent's public score & 39 & incl.\ 3 gamed outright (2 AdderBoard, 1 seven-qubit code) and 5 TESS runs scored by referees later found exploitable\\
Matched its parent, no gain & 58 & incl.\ at least 10 that delivered the parent essentially unchanged, per their own notebooks\\
Worse than its parent & 2 & both TESS, under a stricter referee\\
No valid solution & 5 & 4 stopped by the Modal spend limit (LM), 1 by the step limit (TESS)\\
\midrule
Hit the per-run spend cap (overlaps the rows above) & 5 & introduced after one run cost \$7.15\\
Lost before scoring, not in the records & 3 & 2 killed with the host (C11), 1 lost when its shell died (SABRE)\\
\bottomrule
\end{tabular}
\caption{Outcome of every recorded proposer run in the fourteen problem campaigns (104 runs, 27 sessions; the 18 proposals of the three autonomous runs are reported in Section~\ref{sec:auto-run}, and the nine of the accelerator campaign in Section~\ref{sec:chip}), computed from the kernel's event logs; the notes come from the run notebooks and Section~\ref{sec:gaming}.}
\label{tab:outcomes}
\end{table}

\paragraph{When the agents caught our mistakes.}
Several runs reported errors in the operator's own scaffolding instead of exploiting them, as AIDE$^2$ also observed~\cite{srikanth2026aide2}:
\begin{itemize}[noitemsep]
\item Both SABRE proposers in session~2 reported that the shipped local verifier had unfilled \verb|{BASE}|/\verb|{TARGET}| placeholders and crashed. They worked around it and said so.
\item A SABRE proposer noticed that two benchmark files are byte-identical. Another proved by exhaustive placement that one circuit has no zero-SWAP embedding, so the published count is optimal.
\item A MUB proposer found that the operator's polar retraction gives NaN gradients at the identity basis, which the record contains, and documented the fix.
\item The AdderBoard proposers listed the constants loophole under ``review risk'' in their final message. The TESS proposer stated its intent to exploit the uncapped classes in its notebook.
\end{itemize}
In two of the four gaming incidents, then, the exploit was stated in the agents' own reports before the operator found it in the scores. Reading the notebooks is a cheap detector.

\paragraph{Infrastructure lessons that became kernel features.}
\begin{itemize}[noitemsep]
\item A proposer that ran 40 minutes on 2.1M input tokens cost \$7.15 by itself, which pushed a \$10 session to \$13.88. This led to the per-run spend cap.
\item A proposer that hit the step limit had its work discarded. The kernel now streams results and keeps the last state.
\item A discovery session launched from a foreground shell lost a proposer run unharvested when the shell died. Sessions now run as daemon jobs.
\item A verification sandbox with a hard-coded 2~GB limit OOM-killed a valid $n=7$ qutrit evaluation. Verifier resources are now set per task.
\item Two autonomous jobs launched in the same second shared a session directory. Session names now carry a random suffix.
\item A full chip evaluation takes 10--12 minutes, and the kernel re-scores a parent whenever it is sampled. The verifier's raw output is now cached by verifier, solution, split and image, and the score is recomputed from it on every read, so a corrected score range does not invalidate the cache.
\end{itemize}
Each of these looked minor, and each would have silently cost a result.

\paragraph{Mistakes in the chip evaluator.} The accelerator task needed four fixes from the operator before its scores could be trusted, and two of them repeat earlier lessons.
\begin{itemize}[noitemsep]
\item \textbf{Score range reversed.} The first launch mapped the original design to a score of 1. We cancelled it before any proposer ran. An autonomous scout later made the same mistake (Section~\ref{sec:auto-run}).
\item \textbf{Saturation, again.} In session~1 the score reached 1 at 289~fJ, half the baseline. Both first-round proposers beat that bound in their first run, so later runs had nothing to gain, and an \emph{aggressive rewrite} proposer shipped its parent unchanged. This is the AdderBoard saturation of Section~\ref{sec:gaming} without the gaming. We cancelled the session, set the score-1 point to 100~fJ, and reseeded session~2 with session~1's best design.
\item \textbf{Silent simulation.} Icarus left the internal delayed signals of the ASAP7 sequential cell models undriven, so the first gate-level runs produced no outputs at all. The evaluator now compiles functional versions of the cell models.
\item \textbf{The wrong DRC count.} The first parser read the DRC count from the first routing iteration rather than the last. It now reads the final count by exact key.
\end{itemize}
None of these favoured a candidate, but each would have produced wrong numbers in the tables above if a person had not read the logs.

\paragraph{Where progress came from within a session.}
Figure~\ref{fig:progress} plots every scored proposal on six problems in the order it ran, with the best value so far. Three patterns hold across them.
\begin{itemize}[noitemsep]
\item \textbf{Gains come early in a session.} Every improvement was set in the first or second iteration of its session; no third-iteration run set a new best on any problem.
\item \textbf{A few arms carry the gains.} The runs that set new bests drew \emph{aggressive rewrite} (qutrit twice, Golay, SABRE), \emph{new neighbourhood} (C11, the SABRE beam search, two LM steps), \emph{data structure} (SABRE), \emph{structured construction} (stabilizer rank), and \emph{connectivity} and \emph{parameter sharing} (the LM). \emph{Conservative} set a new best twice (SABRE, LM) and \emph{robust/simple} once (SABRE), only where the seed was far from optimal; on a record seed they never did.
\item \textbf{Most proposals do not improve.} Of 47 scored proposals on these six problems, 16 set a new best. The search is a sequence of rare large steps between plateaus, not a steady climb.
\end{itemize}
The accelerator task, which is not in the figure, fits the first two patterns only in part. Its last and largest step came in the third iteration of session~2, from \emph{new neighbourhood}. A \emph{conservative} run there did useful work, because the notes held a concrete untried combination: it merged the seed's clock gating with a parallel run's arithmetic (274.9 to 252.8~fJ). A second \emph{conservative} draw on the 188.2~fJ design shipped it unchanged.

\paragraph{Cost.}
The fourteen problem campaigns used about \$171 of LLM usage, and about \$180 with the gated evolve run, plus Modal compute. The three autonomous runs cost a further \$27.42 end to end, and the accelerator campaign \$11.0 plus Modal CPU time. Single finding runs cost between \$0.60 (the C11 state) and \$4.15 (the $T^{\otimes7}$ decomposition). Host-side certification was sometimes the expensive part: about 45 CPU-minutes per endpoint for the C11 certificate.

\section{Safety}
\label{sec:safety}
A system that edits its own harness and runs code it wrote needs containment that does not rely on the model's cooperation. The \IM{} enforces the following structurally, in the kernel.
\begin{itemize}[noitemsep]
\item \textbf{Sandboxing.} Every proposer and every scorer runs in a fresh container or Modal sandbox. Network is off by default, the root file system is read-only, and data volumes, where needed, are mounted read-only. Kernel code, private splits, scorers and credentials are never mounted, so an agent cannot read its hidden grade or change how it is graded.
\item \textbf{Immutable grading.} The kernel scores solutions in a separate sandbox and never trusts an agent's report of what it ran. A genome edit that touches kernel paths, or changes a kernel-owned file by a single byte, is rejected before grading.
\item \textbf{Bounded spend.} A per-run cap ends an agent at its next model call once its cost reaches the limit, and a session budget stops new proposers.
\item \textbf{Human gates and their machine replacement.} A self-modification whose gain exceeds the anomaly threshold is parked for review instead of activated. Every generation can be rolled back. Outside autonomous mode, proposed evaluators are inert until a human approves them. In autonomous mode that approval is automatic only after three checks: kernel calibration on known cases, an adversarial audit whose exploits are replayed, and a blind independent re-check of the final result. The approval and its basis are logged. A verifier written by one agent is therefore never trusted on that agent's word alone. In the first of the three autonomous runs, this chain rejected the first task.
\item \textbf{No outward actions.} Agents cannot publish. Every external step in this paper was taken by the operator: the Zenodo deposits, the catalogue issues and pull requests, and the arena posts. Autonomous mode has network access only in the scout and audit phases, to read catalogues and the literature, and it ends with a report, not a submission.
\end{itemize}
What we observed is mostly in Section~\ref{sec:gaming}. Proposers gamed lenient scorers four times, and in two cases said so openly: once when asked for review risks (AdderBoard), and once in a notebook (TESS). We saw no attempt to read hidden splits or to escape a sandbox. However, unlike STOP~\cite{zelikman2023stop}, we did not audit traces systematically for such attempts, so this is an absence of observed incidents, not a measured rate. The main residual risk is not the agent but the harness: our private splits were only as good as the verifiers the operator wrote, and two of them were exploitable for a full iteration before they were fixed.

\section{Related work}
\paragraph{LLM-guided program search.} FunSearch~\cite{romeraparedes2024funsearch} evolved short Python functions scored by an evaluator. AlphaEvolve~\cite{novikov2025alphaevolve} scaled this to whole files, rich prompts and many problems, and Georgiev et al.~\cite{georgiev2025mathematical} applied it to 67 problems in mathematical analysis, geometry and combinatorics. ShinkaEvolve~\cite{lange2025shinka} and Hyra~\cite{hyra2026} are open, sample-efficient variants; Hyra uses agentic proposers and lets the evaluator evolve. AIDE~\cite{jiang2025aide} searches over ML code with a tree of drafts. The \IM{} shares their reliance on automatic evaluation but differs in kind: its proposals are hour-long agent runs rather than sampled programs, it also searches open-endedly in real data, and its own harness is subject to gated self-modification.
\paragraph{Self-improving agents.} STOP~\cite{zelikman2023stop} improved its own scaffolding program. The Darwin G\"odel Machine~\cite{zhang2025dgm} evolved coding agents with an archive and documented reward hacking of its own tests. Hyperagents~\cite{zhang2026hyperagents} made the meta agent editable. AIDE$^2$~\cite{srikanth2026aide2} rewrote a research agent's harness under a private grade it never sees. Prime Agent~\cite{karten2026prime} and Continual Harness~\cite{karten2026continual} keep typed supplemental state with rollback by ID, building on recursive language models~\cite{zhang2025rlm}. ACE~\cite{zhang2025ace} curates an itemised playbook. Weng~\cite{weng2026harness} surveys harness engineering for self-improvement. The \IM{} combines an editable meta agent, an archive, a cost-aware gate and an ACE-style playbook, and adds an honest report that, so far, its discovery results come from the unmodified genome.
\paragraph{Agents for science.} ReAct~\cite{yao2022react} and Reflexion~\cite{shinn2023reflexion} established tool use and verbal feedback. The AI Scientist~\cite{lu2024aiscientist} automated the paper cycle in ML. Our emphasis is narrower and older: a claim counts only if a program that the agent did not write can check it.

\section{Discussion}
\paragraph{What worked.} Giving a strong model a shell, an hour and an exact verifier was enough to move four catalogued open problems and one standard benchmark in six days for under \$200 of model usage. The winning moves were often changes of representation. Two were not suggested anywhere in the operator's notes:
\begin{itemize}[noitemsep]
\item counting mixed qubits instead of non-$Z$ qubits (Golay);
\item shrinking $n$ instead of growing it (qutrit).
\end{itemize}
Two others implemented directions the notes listed as untried: a symmetric dictionary instead of annealing over all stabilizer states (stabilizer rank), and a beam over SWAP sequences instead of a greedy SWAP choice (SABRE).
The public/private split and adversarially calibrated verifiers, not instructions, kept the outputs honest. The same idea carried over to autonomous mode. When an agent writes its own evaluator, the evaluator is trusted only after machine checks that a different agent and the kernel perform. In the first end-to-end run these checks caught a stale record that would otherwise have become a false discovery. The approach also carried over to hardware. With power measured from gate-level switching activity on a routed design, and correctness checked bit-exactly at both RTL and gate level, the proposers cut energy per MAC by 77.7\% through architectural changes (dataflow, storage structure, clock gating, generated arithmetic) rather than tuning.
\paragraph{What did not.} Self-improvement has not yet earned its name in our data. The one self-derived generation was rejected on cost, the most consequential harness changes were the operator's, and cross-session memory ran through hand-edited notes. Where a record is a robust optimum of random-restart search (MUBs, Hadamard, the seven-qubit code), the \IM{} reproduced the record and produced negative evidence, not progress. Proposers sometimes found an object without understanding it, as with the product structure of the qutrit state. All three fully autonomous runs ended in ties with public records. The first scout chose a crowded problem, and the first run's final round was wasted on conservative no-ops. Research arms and narrower goals in runs 2 and 3 produced real searches and near misses, but no improvement, and two proposers still declined to search. In run~2 the machine checks that replace human approval passed a task whose score was reversed. They test whether a verifier accepts the right objects, but not whether the score ranks them the right way round.
\paragraph{Limitations.} All results come from a single operator, a single proposer model and one to three sessions per problem. There are no repeated seeds, ablations or matched-compute comparisons. Several problems were chosen \emph{because} they looked search-shaped, so the success rate is not representative. The quantum results are lower bounds or explicit constructions with stated claim boundaries, not optimality results. The SABRE comparison is not compute-matched. The TESS candidate is unconfirmed. The accelerator result rests on a predictive PDK, one flow, one clock, one synthetic workload and zero-delay gate-level simulation, and it compares against the original RTL rather than a hand-optimised design. End-to-end autonomy has been demonstrated three times, all with null results, and in runs 2 and 3 the operator named the table to search. The fifteen operator-led campaigns that produced the results above were not run in autonomous mode, so none of their results is an autonomous end-to-end discovery.
\paragraph{Next steps.} Next we plan to: run autonomous mode repeatedly with open goals, with a scout that also checks leaderboards and discussion threads before choosing, a validation step that checks the score's direction on calibration cases, and a brief that forbids returning the parent without a search; repeat the accelerator task on a second PDK, at several clocks and workload seeds, with timing-annotated simulation; let the meta agent propose strategy arms and brief templates through the gate; replace operator-written ``already measured'' tables with consolidation output, and measure whether the next session does better; run matched A/B comparisons of gated generations on held-out problem families (imp@$k$~\cite{zhang2026hyperagents}); and add a reviewer round to every catalogue-mode session.

\section*{Use of AI systems}
The discoveries reported here were made by Claude Sonnet~5.5 proposer agents inside the \IM. The harness, task verifiers, seeds, figure scripts and this manuscript were written by the author with the assistance of Claude (Anthropic) as a coding and writing assistant, under the author's direction. The author chose the problems, checked every reported number on the host, and takes responsibility for the content.

\section*{Data availability}
States, circuits, decompositions, routings, certificates, agent notebooks and verification code for the published results are on Zenodo~\cite{zen_c11,zen_qutrit,zen_golay,zen_private,zen_stabrank,zen_sabre,zen_tess}. The accelerator designs, evaluator and notebooks, and the three autonomous session directories, are not yet deposited.
\section*{Ethics statement}
The problems studied are in mathematics, quantum information, compilers, machine learning, astronomy and digital hardware design, and carry little direct potential for harm. The broader concern with self-improving agents is that they may optimise a proxy in unintended ways. We observed exactly this on a small scale (Section~\ref{sec:gaming}) and report every instance. All external claims went out under the author's name after host-side verification, with AI involvement disclosed in each deposit, and none was posted by an agent. The TESS candidate is reported as unconfirmed, and no follow-up request was sent on the agent's behalf.

\section*{Reproducibility statement}
For every published result, the Zenodo deposits contain the solution object, the verifier, the task text given to the agents, the agents' notebooks and search code, and SHA-256 sums. Anyone can re-check the results without our harness. Appendix~\ref{app:sessions} lists every discovery session with its configuration and spend, Appendix~\ref{app:arms} the strategy-arm texts, and Appendix~\ref{app:gate} the gate configuration and the self-derived playbook. For each autonomous run, the session directory keeps every artefact: the goal, each scout round's task, the validation reports, both audits with their exploit files and replay results, the automatic approval record, the search bank, the independent checker's code and verdict, and the final report and claim. Appendix~\ref{app:autorun} summarises it. Model runs are not deterministic: rerunning a session will produce different proposals, and we report only one session sequence per problem. The model versions were Claude Sonnet~5.5 for proposers and Claude Opus~5.5 for the meta agent.

\bibliographystyle{unsrtnat}
\bibliography{refs}

\begin{thebibliography}{49}
\providecommand{\natexlab}[1]{#1}
\providecommand{\url}[1]{\texttt{#1}}
\expandafter\ifx\csname urlstyle\endcsname\relax
  \providecommand{\doi}[1]{doi: #1}\else
  \providecommand{\doi}{doi: \begingroup \urlstyle{rm}\Url}\fi

\bibitem[Novikov et~al.(2025)Novikov, V{\~u}, Eisenberger, Dupont, Huang,
  Wagner, Shirobokov, Kozlovskii, Ruiz, Mehrabian, Kumar, See, Chaudhuri,
  Holland, Davies, Nowozin, Kohli, and Balog]{novikov2025alphaevolve}
Alexander Novikov, Ng{\^a}n V{\~u}, Marvin Eisenberger, Emilien Dupont, Po-Sen
  Huang, Adam~Zsolt Wagner, Sergey Shirobokov, Borislav Kozlovskii, Francisco
  J.~R. Ruiz, Abbas Mehrabian, M.~Pawan Kumar, Abigail See, Swarat Chaudhuri,
  George Holland, Alex Davies, Sebastian Nowozin, Pushmeet Kohli, and Matej
  Balog.
\newblock {AlphaEvolve}: A coding agent for scientific and algorithmic
  discovery.
\newblock arXiv:2506.13131, 2025.

\bibitem[Romera-Paredes et~al.(2024)Romera-Paredes, Barekatain, Novikov, Balog,
  Kumar, Dupont, Ruiz, Ellenberg, Wang, Fawzi, Kohli, and
  Fawzi]{romeraparedes2024funsearch}
Bernardino Romera-Paredes, Mohammadamin Barekatain, Alexander Novikov, Matej
  Balog, M.~Pawan Kumar, Emilien Dupont, Francisco J.~R. Ruiz, Jordan~S.
  Ellenberg, Pengming Wang, Omar Fawzi, Pushmeet Kohli, and Alhussein Fawzi.
\newblock Mathematical discoveries from program search with large language
  models.
\newblock \emph{Nature}, 625:\penalty0 468--475, 2024.
\newblock \doi{10.1038/s41586-023-06924-6}.

\bibitem[Zelikman et~al.(2023)Zelikman, Lorch, Mackey, and
  Kalai]{zelikman2023stop}
Eric Zelikman, Eliana Lorch, Lester Mackey, and Adam~Tauman Kalai.
\newblock Self-taught optimizer ({STOP}): Recursively self-improving code
  generation.
\newblock arXiv:2310.02304, 2023.

\bibitem[Zhang et~al.(2025{\natexlab{a}})Zhang, Hu, Lu, Lange, and
  Clune]{zhang2025dgm}
Jenny Zhang, Shengran Hu, Cong Lu, Robert Lange, and Jeff Clune.
\newblock {Darwin G\"odel Machine}: Open-ended evolution of self-improving
  agents.
\newblock arXiv:2505.22954, 2025{\natexlab{a}}.

\bibitem[Zhang et~al.(2026)Zhang, Zhao, Yang, Foerster, Clune, Jiang, Devlin,
  and Shavrina]{zhang2026hyperagents}
Jenny Zhang, Bingchen Zhao, Wannan Yang, Jakob Foerster, Jeff Clune, Minqi
  Jiang, Sam Devlin, and Tatiana Shavrina.
\newblock Hyperagents.
\newblock arXiv:2603.19461, 2026.

\bibitem[Srikanth et~al.(2026)Srikanth, Zhao, Xu, Wu, and
  Jiang]{srikanth2026aide2}
Dhruv Srikanth, Bingchen Zhao, Dixing Xu, Yuxiang Wu, and Zhengyao Jiang.
\newblock Recursive self-improvement of {AI} research agents.
\newblock arXiv:2609.26457, 2026.

\bibitem[Good(1966)]{good1965}
Irving~John Good.
\newblock Speculations concerning the first ultraintelligent machine.
\newblock In \emph{Advances in Computers}, volume~6, pages 31--88. Academic
  Press, 1966.
\newblock \doi{10.1016/S0065-2458(08)60418-0}.

\bibitem[{LangChain}()]{deepagents}
{LangChain}.
\newblock Deep agents.
\newblock \url{https://docs.langchain.com/oss/python/deepagents/overview}.
\newblock Documentation, accessed October 2026.

\bibitem[Weng(2026)]{weng2026harness}
Lilian Weng.
\newblock Harness engineering for self-improvement.
\newblock \url{https://lilianweng.github.io/posts/2026-07-04-harness/}, 2026.
\newblock Blog post, 4 July 2026.

\bibitem[{Hyra Team}(2026)]{hyra2026}
{Hyra Team}.
\newblock Hyra: A simple yet effective scaffold for general discovery.
\newblock \url{https://hy.tencent.ai/research/hyra}, 2026.
\newblock Tencent Hunyuan blog post, 21 July 2026.

\bibitem[Zhang et~al.(2025{\natexlab{b}})Zhang, Hu, Upasani, Ma, Hong,
  Kamanuru, Rainton, Wu, Ji, Li, Thakker, Zou, and Olukotun]{zhang2025ace}
Qizheng Zhang, Changran Hu, Shubhangi Upasani, Boyuan Ma, Fenglu Hong,
  Vamsidhar Kamanuru, Jay Rainton, Chen Wu, Mengmeng Ji, Hanchen Li, Urmish
  Thakker, James Zou, and Kunle Olukotun.
\newblock Agentic context engineering: Evolving contexts for self-improving
  language models.
\newblock arXiv:2510.04618, 2025{\natexlab{b}}.

\bibitem[Santhanalakshmi~Ganesan(2026{\natexlab{a}})]{zen_c11}
Harish Santhanalakshmi~Ganesan.
\newblock Certified positive coherent information of the qubit depolarizing
  channel at per-pauli $p = 32483/500000$ (45 qubits, rank-2 symmetric input):
  state, exact certificates, evaluator.
\newblock Zenodo, 2026{\natexlab{a}}.

\bibitem[Santhanalakshmi~Ganesan(2026{\natexlab{b}})]{zen_qutrit}
Harish Santhanalakshmi~Ganesan.
\newblock A two-dimensional-subspace input beats the hashing point of the qudit
  depolarizing channel: positive single-letter coherent information at $p =
  0.2910672$ for $d = 3$.
\newblock Zenodo, 2026{\natexlab{b}}.

\bibitem[Santhanalakshmi~Ganesan(2026{\natexlab{c}})]{zen_golay}
Harish Santhanalakshmi~Ganesan.
\newblock A 44-gate, depth-6 {Clifford} circuit preparing the [[23,1,7]]
  {Golay} logical zero: a new point on the {CNOT}-count/depth frontier.
\newblock Zenodo, 2026{\natexlab{c}}.

\bibitem[Santhanalakshmi~Ganesan(2026{\natexlab{d}})]{zen_private}
Harish Santhanalakshmi~Ganesan.
\newblock Private information of the qubit depolarizing channel beyond the
  single-use threshold: explicit finite-$n$ witnesses up to $p = 0.28467$, and
  what one-way key distillation already implies.
\newblock Zenodo, 2026{\natexlab{d}}.

\bibitem[Santhanalakshmi~Ganesan(2026{\natexlab{e}})]{zen_stabrank}
Harish Santhanalakshmi~Ganesan.
\newblock An explicit ten-term stabilizer decomposition of $|t\rangle^{\otimes
  7}$: $\chi(|t\rangle^{\otimes 7}) \le 10$.
\newblock Zenodo, 2026{\natexlab{e}}.

\bibitem[Santhanalakshmi~Ganesan(2026{\natexlab{f}})]{zen_sabre}
Harish Santhanalakshmi~Ganesan.
\newblock Beam search over {SWAP} sequences on the {SABRE} benchmark: 82\%
  fewer added gates than {SABRE} and 29\% fewer than the best published
  routings for {IBM Q20 Tokyo}, found by an autonomous agent.
\newblock Zenodo, 2026{\natexlab{f}}.

\bibitem[Santhanalakshmi~Ganesan(2026{\natexlab{g}})]{zen_tess}
Harish Santhanalakshmi~Ganesan.
\newblock A 1.6 {$R_\oplus$} transiting planet candidate on a 5.31-day orbit
  around the nearby {M4.5} dwarf {G 249-11} ({TIC 417732194}).
\newblock Zenodo, 2026{\natexlab{g}}.

\bibitem[Fern and Whaley(2008)]{fern2008}
Jesse Fern and K.~Birgitta Whaley.
\newblock Lower bounds on the nonzero capacity of {Pauli} channels.
\newblock \emph{Physical Review A}, 78:\penalty0 062335, 2008.
\newblock \doi{10.1103/PhysRevA.78.062335}.

\bibitem[Agarwal et~al.(2026)Agarwal, Kalra, Lee, Leung, Schaeffer, Sinha, and
  Smith]{agarwal2026}
Avantika Agarwal, Amolak~Ratan Kalra, Sungjai Lee, Debbie Leung, Luke
  Schaeffer, Pulkit Sinha, and Graeme Smith.
\newblock Enhanced quantum capacity thresholds from symmetry.
\newblock arXiv:2605.09138, 2026.

\bibitem[Krohn-Grimberghe(2026)]{krohngrimberghe2026}
Artus Krohn-Grimberghe.
\newblock A certified lower bound on the quantum-capacity threshold of the
  depolarizing channel.
\newblock arXiv:2608.15870, 2026.

\bibitem[Renes and Smith(2007)]{renes2007}
Joseph~M. Renes and Graeme Smith.
\newblock Noisy processing and distillation of private quantum states.
\newblock \emph{Physical Review Letters}, 98:\penalty0 020502, 2007.
\newblock \doi{10.1103/PhysRevLett.98.020502}.

\bibitem[Kern and Renes(2008)]{kern2008}
Oliver Kern and Joseph~M. Renes.
\newblock Improved one-way rates for {BB84} and 6-state protocols.
\newblock \emph{Quantum Information and Computation}, 8\penalty0
  (8--9):\penalty0 756--772, 2008.
\newblock \doi{10.26421/QIC8.8-9-6}.

\bibitem[Doherty et~al.(2026)Doherty, Puviani, Brewer, Matos, Amaro, Criger,
  and Stephen]{doherty2026golay}
Michael Doherty, Matteo Puviani, Jasmine Brewer, Gabriel Matos, David Amaro,
  Ben Criger, and David~T. Stephen.
\newblock Fast stabilizer state preparation via {AI}-optimized graph
  decimation.
\newblock arXiv:2603.17743, 2026.

\bibitem[Bravyi et~al.(2016)Bravyi, Smith, and Smolin]{bravyi2016}
Sergey Bravyi, Graeme Smith, and John~A. Smolin.
\newblock Trading classical and quantum computational resources.
\newblock \emph{Physical Review X}, 6:\penalty0 021043, 2016.
\newblock \doi{10.1103/PhysRevX.6.021043}.

\bibitem[Qassim et~al.(2021)Qassim, Pashayan, and Gosset]{qassim2021}
Hammam Qassim, Hakop Pashayan, and David Gosset.
\newblock Improved upper bounds on the stabilizer rank of magic states.
\newblock \emph{Quantum}, 5:\penalty0 606, 2021.
\newblock \doi{10.22331/q-2021-12-20-606}.

\bibitem[Li et~al.(2019)Li, Ding, and Xie]{li2019sabre}
Gushu Li, Yufei Ding, and Yuan Xie.
\newblock Tackling the qubit mapping problem for {NISQ}-era quantum devices.
\newblock In \emph{Proceedings of the 24th International Conference on
  Architectural Support for Programming Languages and Operating Systems
  (ASPLOS)}, pages 1001--1014, 2019.
\newblock \doi{10.1145/3297858.3304023}.

\bibitem[Xu et~al.(2025)Xu, Li, Liu, Li, and Wang]{xu2025hail}
Kang Xu, Zeyang Li, Xinjian Liu, Dandan Li, and Yukun Wang.
\newblock An efficient iterative algorithm for qubit mapping via layer-weight
  assignment and search space reduction.
\newblock arXiv:2502.07536, 2025.

\bibitem[add()]{adderboard}
{AdderBoard}: the smallest transformer that adds two 10-digit numbers.
\newblock \url{https://github.com/anadim/AdderBoard}.
\newblock Leaderboard repository, accessed October 2026.

\bibitem[Jordan et~al.()]{moddednanogpt}
Keller Jordan et~al.
\newblock modded-nanogpt.
\newblock \url{https://github.com/KellerJordan/modded-nanogpt}.
\newblock Accessed October 2026.

\bibitem[Ajayi et~al.(2019)Ajayi, Chhabria, Foga{\c{c}}a, Hashemi, Hosny,
  Kahng, Kim, Lee, Mallappa, Neseem, Pradipta, Reda, Saligane, Sapatnekar,
  Sechen, Shalan, Swartz, Wang, Wang, Woo, and Xu]{orfs}
Tutu Ajayi, Vidya~A. Chhabria, Mateus Foga{\c{c}}a, Soheil Hashemi, Abdelrahman
  Hosny, Andrew~B. Kahng, Minsoo Kim, Jeongsup Lee, Uday Mallappa, Marina
  Neseem, Geraldo Pradipta, Sherief Reda, Mehdi Saligane, Sachin~S. Sapatnekar,
  Carl Sechen, Mohamed Shalan, William Swartz, Lutong Wang, Zhehong Wang,
  Mingyu Woo, and Bangqi Xu.
\newblock Toward an open-source digital flow: First learnings from the
  {OpenROAD} project.
\newblock In \emph{Proceedings of the 56th Annual Design Automation Conference
  (DAC)}, pages 1--4, 2019.
\newblock \doi{10.1145/3316781.3326334}.

\bibitem[Clark et~al.(2016)Clark, Vashishtha, Shifren, Gujja, Sinha, Cline,
  Ramamurthy, and Yeric]{asap7}
Lawrence~T. Clark, Vinay Vashishtha, Lucian Shifren, Aditya Gujja, Saurabh
  Sinha, Brian Cline, Chandarasekaran Ramamurthy, and Greg Yeric.
\newblock {ASAP7}: A 7-nm {finFET} predictive process design kit.
\newblock \emph{Microelectronics Journal}, 53:\penalty0 105--115, 2016.
\newblock \doi{10.1016/j.mejo.2016.04.006}.

\bibitem[Ricker et~al.(2015)Ricker, Winn, Vanderspek, Latham,
  et~al.]{ricker2015}
George~R. Ricker, Joshua~N. Winn, Roland Vanderspek, David~W. Latham, et~al.
\newblock Transiting exoplanet survey satellite.
\newblock \emph{Journal of Astronomical Telescopes, Instruments, and Systems},
  1:\penalty0 014003, 2015.
\newblock \doi{10.1117/1.JATIS.1.1.014003}.

\bibitem[ein()]{einsteinarena}
Einstein arena.
\newblock \url{https://einsteinarena.com}.
\newblock Accessed October 2026.

\bibitem[Butbaia et~al.(2026)Butbaia, Tan, Vipulanandan, Huang,
  Saunders-A'Court, Fagan, Passaro, Tarquini, and Gukov]{hadamard2026}
Giorgi Butbaia, Justin Tan, Pragatheeswaran Vipulanandan, Xiaoyu Huang, Toby
  Saunders-A'Court, Lucas Fagan, Davide Passaro, Michele Tarquini, and Sergei
  Gukov.
\newblock New records for the hadamard maximal determinant problem.
\newblock arXiv:2608.22518, 2026.

\bibitem[Rains(1997)]{rains1997}
Eric~M. Rains.
\newblock Quantum codes of minimum distance two.
\newblock arXiv:quant-ph/9704043, 1997.

\bibitem[Angl{\`e}s~Munn{\'e} and Huber(2026)]{anglesmunne2026}
Gerard Angl{\`e}s~Munn{\'e} and Felix Huber.
\newblock {SDP} bounds on quantum codes: rational certificates.
\newblock arXiv:2603.19901, 2026.

\bibitem[Raynal et~al.(2011)Raynal, L{\"u}, and Englert]{raynal2011}
Philippe Raynal, Xin L{\"u}, and Berthold-Georg Englert.
\newblock Mutually unbiased bases in dimension six: The four most distant
  bases.
\newblock \emph{Physical Review A}, 83:\penalty0 062303, 2011.
\newblock \doi{10.1103/PhysRevA.83.062303}.

\bibitem[Georgiev et~al.(2025)Georgiev, G{\'o}mez-Serrano, Tao, and
  Wagner]{georgiev2025mathematical}
Bogdan Georgiev, Javier G{\'o}mez-Serrano, Terence Tao, and Adam~Zsolt Wagner.
\newblock Mathematical exploration and discovery at scale.
\newblock arXiv:2511.02864, 2025.

\bibitem[Lange et~al.(2025)Lange, Imajuku, and Cetin]{lange2025shinka}
Robert~Tjarko Lange, Yuki Imajuku, and Edoardo Cetin.
\newblock {ShinkaEvolve}: Towards open-ended and sample-efficient program
  evolution.
\newblock arXiv:2509.19349, 2025.

\bibitem[Jiang et~al.(2025)Jiang, Schmidt, Srikanth, Xu, Kaplan, Jacenko, and
  Wu]{jiang2025aide}
Zhengyao Jiang, Dominik Schmidt, Dhruv Srikanth, Dixing Xu, Ian Kaplan, Deniss
  Jacenko, and Yuxiang Wu.
\newblock {AIDE}: {AI}-driven exploration in the space of code.
\newblock arXiv:2502.13138, 2025.

\bibitem[Karten et~al.(2026{\natexlab{a}})Karten, Zhang, Thomas, M{\"u}ller,
  Bakouch, Auras, Senghaas, Obeid, Dunas, Hagemann, and
  Jaghouar]{karten2026prime}
Seth Karten, Alex~L. Zhang, Kevin Thomas, Sebastian M{\"u}ller, Elie Bakouch,
  Daniel Auras, Mika Senghaas, Fares Obeid, Konstantin Dunas, Johannes
  Hagemann, and Sami Jaghouar.
\newblock Prime agent: A self-improving {RLM} harness.
\newblock arXiv:2608.23552, 2026{\natexlab{a}}.

\bibitem[Karten et~al.(2026{\natexlab{b}})Karten, Zhang, Upaa, Feng, Li, Shi,
  Jin, and Vodrahalli]{karten2026continual}
Seth Karten, Joel Zhang, Tersoo Upaa, Ruirong Feng, Wenzhe Li, Chengshuai Shi,
  Chi Jin, and Kiran Vodrahalli.
\newblock Continual harness: Online adaptation for self-improving foundation
  agents.
\newblock arXiv:2605.09998, 2026{\natexlab{b}}.

\bibitem[Zhang et~al.(2025{\natexlab{c}})Zhang, Kraska, and
  Khattab]{zhang2025rlm}
Alex~L. Zhang, Tim Kraska, and Omar Khattab.
\newblock Recursive language models.
\newblock arXiv:2512.24601, 2025{\natexlab{c}}.

\bibitem[Yao et~al.(2022)Yao, Zhao, Yu, Du, Shafran, Narasimhan, and
  Cao]{yao2022react}
Shunyu Yao, Jeffrey Zhao, Dian Yu, Nan Du, Izhak Shafran, Karthik Narasimhan,
  and Yuan Cao.
\newblock {ReAct}: Synergizing reasoning and acting in language models.
\newblock arXiv:2210.03629, 2022.

\bibitem[Shinn et~al.(2023)Shinn, Cassano, Berman, Gopinath, Narasimhan, and
  Yao]{shinn2023reflexion}
Noah Shinn, Federico Cassano, Edward Berman, Ashwin Gopinath, Karthik
  Narasimhan, and Shunyu Yao.
\newblock Reflexion: Language agents with verbal reinforcement learning.
\newblock arXiv:2303.11366, 2023.

\bibitem[Lu et~al.(2024)Lu, Lu, Lange, Foerster, Clune, and
  Ha]{lu2024aiscientist}
Chris Lu, Cong Lu, Robert~Tjarko Lange, Jakob Foerster, Jeff Clune, and David
  Ha.
\newblock The {AI} scientist: Towards fully automated open-ended scientific
  discovery.
\newblock arXiv:2408.06292, 2024.

\bibitem[Bruneton et~al.()]{oqp}
Jean-Philippe Bruneton et~al.
\newblock Open quantum problems.
\newblock \url{https://github.com/jpbruneton/open_quantum_problems}.
\newblock Catalogue behind openquantumproblems.com, accessed October 2026.

\bibitem[Guo et~al.()]{qiqczoo}
Naixu Guo et~al.
\newblock {QIQC} open problem zoo.
\newblock \url{https://github.com/Naixu-Guo/quantum-open-problems}.
\newblock Accessed October 2026.

\end{thebibliography}

\clearpage
\appendix

\section{Details of each discovery}
\label{app:discoveries}
\sloppy
For every result in Table~\ref{tab:results} we give the definition, the known results, the result, how it was found and how it was verified. ``How it was found'' identifies the run that produced the object: session and run identifiers, start time (UTC, October 2026), strategy arm, model, wall time, tokens and model cost, all from the kernel's run records. It then gives the method with its parameters, as recorded in the proposer's notebook, and the search effort, including what was tried without success. ``Verification'' lists each check separately: what was checked, by which program and by whom, the acceptance criterion and the outcome, followed by a command that reproduces it from the Zenodo deposit. All proposers are Claude Sonnet~5.5.

\subsection{Quantum capacity of the qubit depolarizing channel}
Let $\mathcal D_p(\rho)=(1-3p)\rho+p(X\rho X+Y\rho Y+Z\rho Z)$, with $p$ the per-Pauli error probability. By the Lloyd--Shor--Devetak theorem, $Q(\mathcal D_p)>0$ whenever some input $\rho$ on $n$ channel uses has positive coherent information $I_c(\rho,\mathcal D_p^{\otimes n})=S(\mathcal D_p^{\otimes n}(\rho))-S((\mathcal D_p^{\otimes n}\otimes\mathrm{id})(\psi_\rho))$, where $\psi_\rho$ purifies $\rho$. The threshold $p^*=\sup\{p: Q(\mathcal D_p)>0\}$ is unknown (catalogue entry C11~\cite{oqp}). Known lower bounds include $0.0637675$~\cite{fern2008} and $0.064912$~\cite{agarwal2026}. The best certified point was $p=16239/250000=0.064956$, for a 45-copy rank-two input supported on the symmetric subspace, certified in exact rational arithmetic~\cite{krohngrimberghe2026}.

\paragraph{Result.} A 45-qubit rank-two input on the symmetric subspace whose coherent information is certified positive at $p=32483/500000=0.064966$, which raises the certified point by $1.0\times10^{-5}$. Its fidelity with the previous record state is $0.99927$.

\paragraph{How it was found.}
\begin{itemize}[noitemsep,leftmargin=*]
\item \textbf{Run.} Session \texttt{20261019-181124}, iteration~0, proposer~0. Started 18:11; \emph{new neighbourhood} arm; 2,522~s; 181,073 input and 3,457 output tokens; \$0.60.
\item \textbf{Start.} The record state A, given as the seed (kernel score 0.478). The tools were the operator's float evaluator, which reduces $I_c$ for symmetric-subspace inputs to Schur--Weyl blocks via a coherent-state generating function and never forms a $2^n$-dimensional matrix, and its autograd twin.
\item \textbf{Method.} The proposer wrote its own checkpointing Adam polish on the real and imaginary parts of the two unnormalised Dicke-basis rows (learning rate $5\times10^{-5}$). It worked at a fixed $p$ and raised $p$ by $10^{-6}$ whenever $I_c>2\times10^{-9}$, for about 5,600 steps, writing the current state to \texttt{/work/solution} throughout.
\item \textbf{Delivered.} A claimed $p=0.064967$ and kernel score 0.4835.
\item \textbf{Also tried.} A second iteration (\$0.72) tried a third row of norm 0.03 (rank three) and a Dicke lift to $n=46$. Neither beat the delivered state.
\end{itemize}

\paragraph{Verification.}
\begin{enumerate}[noitemsep,leftmargin=*]
\item \textbf{Kernel.} The task verifier ran in a fresh sandbox. It accepts a claimed $p$ only if $I_c>10^{-11}$ in two independent quadrature bases (Fibonacci-sphere points). Outcome: accepted at $p=0.064967$.
\item \textbf{Float scan (operator).} $I_c=+2.63\times10^{-7}$, $+1.21\times10^{-7}$, $+2.58\times10^{-8}$, $+2.05\times10^{-9}$ and $-2.17\times10^{-8}$ bits at $p=0.064956$, $0.064962$, $0.064966$, $0.064967$ and $0.064968$. The float threshold is therefore $\approx0.0649671$.
\item \textbf{Exact certificate.} The record authors' certifier (\texttt{certified\_coherent\_information}; rational arithmetic, 224-bit square-root enclosures, 50-digit directed accumulation) was run unchanged on the host, taking 8,407~s and 8,361~s:
\begin{center}\small
\begin{tabular}{lll}
\toprule
$p$ (exact) & certified enclosure of $I_c$ (bits) & float evaluator\\
\midrule
$32481/500000=0.064962$ & $[1.1807,\,1.2256]\times10^{-7}$ & $1.2062\times10^{-7}$\\
$32483/500000=0.064966$ & $[2.3084,\,2.7775]\times10^{-8}$ & $2.5763\times10^{-8}$\\
\bottomrule
\end{tabular}
\end{center}
Both endpoints are classified certified-positive, and each float value lies inside its enclosure. Since $I_c$ is nonincreasing in $p$ for a fixed input, $Q(\mathcal D_p)>0$ for all $p\le0.064966$.
\item \textbf{Reproduce.} The state is \texttt{payloads/EVO1.c128} (complex128, shape $2\times46$, 1,472 bytes, SHA-256 \texttt{7f11f2f1\ldots 67c5ae}). Run \texttt{python certify.py ../payloads/EVO1.c128 0.064966 out.json} (about 45 CPU-minutes)~\cite{zen_c11}.
\end{enumerate}

\subsection{Positivity frontier of the qutrit depolarizing channel}
Let $\mathcal N_{d,p}(\rho)=(1-p)\rho+p\,I/d$. For $d\ge3$ the catalogue (QIQC zoo~\cite{qiqczoo}) listed only the hashing point, i.e.\ the zero of $I_c$ for the maximally mixed input, which for $d=3$ is $p_h=0.287087$.

\paragraph{Result.} With $\eta(x)=-x\log_2x$ and $\sigma_2$ the maximally mixed state on a two-dimensional subspace of $\mathbb C^d$,
\[
I_c(\sigma_2,\mathcal N_{d,p})=2\,\eta\!\Big(\tfrac{1-p}{2}+\tfrac pd\Big)+(d-2)\,\eta\!\Big(\tfrac pd\Big)-\eta\!\Big(1-p+\tfrac p{2d}\Big)-(2d-1)\,\eta\!\Big(\tfrac p{2d}\Big),
\]
whose zero for $d=3$ lies in $[0.29106723603359969,\ 0.29106723603359988]$. This is above the hashing point by $3.98\times10^{-3}$. The same single-letter gain holds for every $3\le d\le64$ that we checked, for example $d=4$: $0.3170689$ against $0.3089806$.

\paragraph{How it was found.}
\begin{itemize}[noitemsep,leftmargin=*]
\item \textbf{Session~1 run.} Session \texttt{20261020-025047}, iteration~0, proposer~0. Started 03:01; \emph{aggressive rewrite}; 1,934~s; 993,698 input and 19,509 output tokens; \$3.27.
  \begin{itemize}[noitemsep]
  \item \emph{Start:} the seed, a rank-3 repetition input on $n=7$ (threshold $0.287731$), with the operator's exact brute-force evaluator ($n\le7$, rank $\le4$) and its autograd twin.
  \item \emph{Method:} a rank-4 input on $n=4$ built from the repetition rows plus a small noisy fourth row. It ran Adam at fixed $p=0.2874$, taking $I_c$ from $-5\times10^{-4}$ to $+9.9\times10^{-3}$, then a warm-started $p$-ladder.
  \item \emph{Outcome:} threshold $0.289205$, with two seeds agreeing to $0.2892$--$0.2893$.
  \end{itemize}
\item \textbf{Session~2 run (finding run).} Session \texttt{20261020-040144}, iteration~0, proposer~0. Started 04:01; \emph{aggressive rewrite}; 1,726~s; 715,424 input and 14,517 output tokens; \$2.36.
  \begin{itemize}[noitemsep]
  \item \emph{Start:} the session-1 state.
  \item \emph{Observation recorded in the notebook:} with rank 4 allowed, a smaller $n$ is better.
  \item \emph{Method:} Adam $p$-ladders from random starts gave $n=3$: $0.28995$ and $n=2$: $0.29106$. L-BFGS with a strong-Wolfe line search and a bisection ladder on $p$ then reached $0.2910672$, with bisection to $\approx4\times10^{-8}$.
  \item \emph{Global check:} 25 random L-BFGS starts at $n=2$ all converge to the same optimum, with $I_c(0.29108)=-6.5466\times10^{-5}$.
  \item \emph{Also tried:} lifts to $n=3,4$ return to the $n=2$ value.
  \end{itemize}
\item \textbf{Operator.} Computed the reduced states, found the product structure $\sigma_2\otimes\sigma_2$ and derived the closed form above. Two later proposer runs had observed that the state is $P/4$ on a subspace, but neither computed the reduced states.
\end{itemize}

\paragraph{Verification.}
\begin{enumerate}[noitemsep,leftmargin=*]
\item \textbf{Kernel.} The exact brute-force verifier, run in a fresh sandbox, accepts if $I_c>10^{-10}$ at the claimed $p$. Outcome: value $0.2910672$, score 0.81344.
\item \textbf{Host bisection.} Bisection to $10^{-8}$ gives $0.29106723$, with $I_c(0.2910672)=+1.848181\times10^{-7}$ and $I_c(0.291068)=-3.92\times10^{-6}$ bits.
\item \textbf{High precision.} A 30-digit \texttt{mpmath} recomputation at the stored amplitudes gives $1.84818087\times10^{-7}$ bits. The smallest joint eigenvalue is $2.4\times10^{-3}$, so the entropies are smooth there.
\item \textbf{Structure.} $\lVert\rho-\rho_A\otimes\rho_B\rVert=3\times10^{-9}$, and both reduced spectra are $(\tfrac12,\tfrac12,0)$. The closed form agrees with the evaluator.
\item \textbf{Interval certificate.} \texttt{mpmath.iv} at 50 digits gives the enclosure stated above.
\item \textbf{Reproduce.} \texttt{python rank2\_certificate.py}, and \texttt{python verification/highprec\_check.py states/s2-it00-p0\_n2r4\_p0.2910672.json 0.2910672 30} (state SHA-256 \texttt{4c3802af\ldots})~\cite{zen_qutrit}.
\end{enumerate}
The result is elementary enough that it may be known, and no priority is claimed.

\subsection{Private information of the qubit depolarizing channel}
For $\Lambda_p(\rho)=(1-p)\rho+p\,I/2$, the zoo record certified $P(\Lambda_p)>0$ only for $p\le0.259824$. The single-use six-state ensemble with noisy preprocessing~\cite{renes2007} has positive private information below $p=0.28223$.

\paragraph{Result.} Explicit two-member classical ensembles on $n$ qubits with positive private information $I(X;B)-I(X;E)$ at $p=0.28336, 0.28372, 0.28405, 0.28437, 0.28466$ for $n=4,\dots,8$. At $n=8$, member~0 is $\rho_0=\sum_s w_s\ket s\bra s$ with $w_s\propto e^{f_{|s|}}$, where the logits over Hamming weight $0,\dots,8$ are
\[
f=(0.9295,\,-0.8515,\,-1.8970,\,-2.2646,\,-2.4467,\,-2.4550,\,-2.1969,\,-1.1514,\,0.6337).
\]
Member~1 is its bit complement, and the two members have equal priors.

\paragraph{How it was found.}
\begin{itemize}[noitemsep,leftmargin=*]
\item \textbf{Run.} Session \texttt{20261020-201046}, iteration~0, proposer~1. Started 20:11; \emph{new neighbourhood}; 2,635~s; 630,192 input and 14,966 output tokens; \$2.12.
\item \textbf{Start.} The session-1 $n=5$ ensemble (threshold $0.283722$), found by Adam on all $2^n$ logits.
\item \textbf{Method.} The proposer restricted the ensemble to a classical complement pair whose weights are a softmax over $n+1$ Hamming-weight logits. This reduces the environment entropy to $n+1$ eigenproblems. It then ran Adam in \texttt{torch} on the exact classical private information, cross-checked against the verifier, with continuation in $p$ and a final bisection.
\item \textbf{Outcome.} Thresholds $0.28336$, $0.28372$, $0.28406$, $0.28437$ and $0.28467$ for $n=4,\dots,8$.
\item \textbf{Also tried.} Three, four, six and eight classical members at $n=3,4$ (no gain). General complex members at $n=2,3$, started at the classical optimum, do not exceed it, and random starts collapse to zero private information.
\end{itemize}

\paragraph{Verification.}
\begin{enumerate}[noitemsep,leftmargin=*]
\item \textbf{Kernel.} The verifier accepts if the private information is at least $10^{-9}$ bits at the claimed $p$. The evaluator self-tests reproduce the hashing rate, brute-force Stinespring results for $n\le3$, and the 14.11\% Renes--Smith rate.
\item \textbf{Host.} Thresholds by bisection in double precision; at $n=8$ the threshold is $0.2846707$, and the private information at $p=0.28466$ is $2.5\times10^{-7}$ bits.
\item \textbf{High precision.} A 30-digit \texttt{mpmath} recomputation was made for every deposited ensemble ($n=8$ takes about 10 minutes).
\item \textbf{Reproduce.} \texttt{python verify/private\_info.py ensembles/n8\_shell\_symmetric.json} and \texttt{python verify/certify\_mp.py ensembles/n5\_shell\_symmetric.json 30}~\cite{zen_private}.
\end{enumerate}
The literature already implies $P>0$ up to $p\approx0.2915$; that is the operator's argument, given in the deposit. These ensembles are new only as explicit finite-$n$ witnesses.

\subsection{Preparing the Golay logical zero}
The task is to prepare $\ket{0_G}=2^{-11/2}\sum_{c\in C^\perp}\ket c$ of the $[[23,1,7]]$ Golay code from $\ket0^{\otimes23}$ with Clifford gates, no ancillas and free single-qubit gates, minimising the two-qubit gate count $N$ and the two-qubit depth $D$. The published Pareto points were $(N,D)=(45,7)$ and $(44,8)$ from RL plus MCTS graph decimation~\cite{doherty2026golay}, and earlier $(57,7)$, $(56,17)$ and $(45,10)$.

\paragraph{Result.} A circuit with $N=44$ controlled-$Z$ gates in $D=6$ layers, which dominates both published points. In the cyclic coordinates of the Golay code, its two-qubit layers are
\begin{align*}
L_1&:\ (4,10),(17,15),(20,16),(1,8),(13,21),(18,6),(12,11),(5,3),(0,14)\\
L_2&:\ (2,16),(8,10),(17,6),(1,12),(11,21),(15,0),(18,13),(5,22),(9,3),(7,20)\\
L_3&:\ (8,19),(12,17),(11,0),(14,22),(6,4),(3,13),(7,21)\\
L_4&:\ (12,5),(19,16),(0,2),(22,8),(4,21),(3,20)\\
L_5&:\ (16,12),(10,0),(18,19),(6,2),(20,11),(4,9),(21,22)\\
L_6&:\ (5,10),(4,11),(9,16),(2,21),(3,6).
\end{align*}
The interleaved single-qubit Cliffords are in the deposit~\cite{zen_golay}.

\paragraph{How it was found.}
\begin{itemize}[noitemsep,leftmargin=*]
\item \textbf{Run.} Session \texttt{20261020-163836}, iteration~1, proposer~1. Started 17:03; \emph{aggressive rewrite}; 1,491~s; 815,918 input and 27,297 output tokens; \$2.86.
\item \textbf{Start.} The operator's transcription of the published $(44,8)$ circuit.
\item \textbf{Representation.} The circuit is written as 44 controlled-Pauli gates $\mathrm{CP}(a,b,P,Q)$ in a global Pauli frame, so all single-qubit Cliffords are absorbed. A stabilizer tableau stored as 46 32-bit columns, written in C, evaluates about $5\times10^6$ full circuits per second per core (the previous Python version ran at 49,000).
\item \textbf{Cost.} The number of qubits left mixed when the circuit is applied in reverse to the target. A qubit is pure if and only if its $X$- and $Z$-columns are linearly dependent, so cost 0 means a product state in any local basis. The previous session's cost accepted only $Z$-basis product states.
\item \textbf{Search.} A Metropolis walk at a fixed 44 gates ($T=0.3$) with energy $2\cdot\text{cost}+\#\{\text{gates in layers beyond the target depth}\}$. Its moves change a gate's Pauli axes or endpoints, shift it in the order, replace it, or delete one gate and insert another. From the start circuit it reaches depth $\le7$ in about 1~s and depth $\le6$ in 10--60~s. A converter turns the result back into H, S and CZ gates with Pauli sign fixes.
\item \textbf{Not found.} Depth~5: in four runs of 300~s, the best is one gate over the layer limit. 43 gates:
  \begin{itemize}[noitemsep]
  \item an exhaustive delete-two/insert-one scan of 92 million neighbours of the start circuit has minimum cost 2;
  \item four fixed-length-43 annealing runs of 540~s each, about $10^{10}$ evaluations in total, end at cost 2, i.e.\ one Bell pair left.
  \end{itemize}
  Total compute was about 25 CPU-minutes.
\end{itemize}

\paragraph{Verification.}
\begin{enumerate}[noitemsep,leftmargin=*]
\item \textbf{Kernel.} The verifier (\texttt{golay\_core.check\_circuit}) ran in a fresh sandbox and simulates the circuit as a stabilizer tableau. The output group is CSS, with pure-$X$ rank 11 and pure-$Z$ rank 12; all signs are $+$; and the $Z$-code has minimum distance 7, which makes it the Golay code by uniqueness. It reported $N=44$ and earliest-possible (ASAP) depth 6; value $44.6$.
\item \textbf{Relabelling (operator).} \texttt{find\_relabeling.py} finds the wire permutation $\pi$ to the cyclic coordinates. It is the same permutation as for the published start circuit, and the relabelled circuit prepares the cyclic-code state exactly.
\item \textbf{Statevector.} A full $2^{23}$-amplitude simulation gives overlap $1.0$ with $\ket{0_G}$ in cyclic coordinates.
\item \textbf{Reproduce.} \texttt{python verify/golay\_core.py circuits/evo\_44\_depth6\_cyclic.json --sv} (a few seconds; about 130~MB of memory).
\end{enumerate}

\subsection{Stabilizer rank of $\ket T^{\otimes7}$}
For $\ket T=(\ket0+e^{i\pi/4}\ket1)/\sqrt2$, the stabilizer rank $\chi(\psi)$ is the least number of stabilizer states whose span contains $\psi$. The exact values $\chi(\ket T^{\otimes m})$ are known for $m\le4$ ($2,2,3,4$)~\cite{bravyi2016}. For $m=5,6$ the bounds were $6$~\cite{qassim2021}, and for $m=7$ only the product bound $12$.

\paragraph{Result.} With $\zeta=e^{i\pi/4}$, $\ket T^{\otimes7}=2^{-7/2}\sum_{j=1}^{10}c_j\ket{\varphi_j}$ for the ten stabilizer states below. They are written in Dehaene--De Moor form $\ket\varphi=\sum_{y\in\mathbb F_2^k}i^{e(y)}\ket{x_0+Gy}$, with $e(y)=\sum_ia_iy_i+2\sum_{(i,l)\in B}y_iy_l$, where the columns of $G$ are the vectors $g_1,\dots,g_k$. Hence $\chi(\ket T^{\otimes7})\le10$, and by sub-multiplicativity $\chi(\ket T^{\otimes13})\le60$ (previously 72).
\begin{center}\scriptsize
\setlength{\tabcolsep}{3pt}
\begin{tabular}{@{}c c l >{\raggedright\arraybackslash}p{3.3cm} l >{\raggedright\arraybackslash}p{1.9cm} l@{}}
\toprule
$j$ & $k$ & $x_0$ & $g_1,\dots,g_k$ & $a$ & $B$ & $c_j$\\
\midrule
1 & 1 & 0011111 & 1111111 & (2) & -- & $-2\zeta-2\zeta^2$ \\
2 & 4 & 0000001 & 0010001, 0001001, 0000101, 0000011 & (0,0,0,0) & -- & $\zeta$ \\
3 & 4 & 1100000 & 0010001, 0001001, 0000101, 0000011 & (0,0,0,0) & -- & $-1+\zeta-\zeta^3$ \\
4 & 3 & 0000000 & 1000000, 0100000, 0011111 & (2,2,1) & -- & $1-\zeta$ \\
5 & 3 & 0000000 & 1000000, 0100000, 0011111 & (0,0,1) & all pairs & $1+\zeta$ \\
6 & 7 & 0000000 & $I_7$ & (0,0,1,1,1,1,1) & -- & $\tfrac14(1-\zeta-\zeta^2-\zeta^3)$ \\
7 & 7 & 0000000 & $I_7$ & (0,0,0,0,0,0,0) & -- & $\tfrac12(-\zeta+\zeta^3)$ \\
8 & 7 & 0000000 & $I_7$ & (0,0,1,1,1,1,1) & $(1,l),l>1$; $(2,l),l>2$ & $\tfrac14(-1-\zeta+\zeta^2-\zeta^3)$ \\
9 & 6 & 0000000 & 1000001, 0100001, 0010001, 0001001, 0000101, 0000011 & (0,0,0,0,0,0) & (1,2) & $\tfrac12(-1+\zeta+\zeta^2-\zeta^3)$ \\
10 & 5 & 0000000 & 1100001, 0010001, 0001001, 0000101, 0000011 & (0,2,2,2,2) & all pairs & $\tfrac12(-1+\zeta-\zeta^2+\zeta^3)$ \\
\bottomrule
\end{tabular}
\end{center}

\paragraph{How it was found.}
\begin{itemize}[noitemsep,leftmargin=*]
\item \textbf{Run.} Session \texttt{20261021-033358}, iteration~0, proposer~0. Started 03:34; \emph{structured construction}; 1,603~s; 1,216,380 input and 33,582 output tokens; \$4.15. The run stopped at the \$4 per-run cap.
\item \textbf{Dictionary.} The qubits are split into blocks $[a,b]$ with $a+b=7$. Three states (the all-zero state, a block GHZ state and the global GHZ state) are closed under all $S_a\times S_b$-symmetric Cliffords: block-wise H, S and X, CZ within and between blocks, and block-parity CNOT fans. This gives 396--540 vectors, which span the $(a+1)(b+1)$-dimensional symmetric space for $[2,5]$, $[3,4]$ and other partitions, but not for $[1,6]$ or the fully symmetric case.
\item \textbf{Subset search.} Batched simulated annealing over $r$-subsets of the dictionary, minimising the squared least-squares residual of $\ket T^{\otimes7}$, with 1,024 chains at about $10^6$ evaluations per minute per CPU. It reached $r=12$ on $[3,4]$ in 40~s, $r=11$ on $[3,4]$ in 220~s and on $[2,5]$ in 130~s, and then $r=10$ on $[2,5]$. Solutions were converted to exact normal form and checked with the exact verifier.
\item \textbf{Record note.} The header of the run's notebook still reports $r=11$; it was written before $r=10$ was found. The delivered \texttt{solution.json} contains 10 states.
\item \textbf{Also tried.} An exhaustive $\mathbb Z_5$-orbit test over all 2.4 million five-qubit stabilizer states found no single cyclic orbit of five states containing $\ket T^{\otimes5}$. Separately, a companion \emph{exhaustive small case} run stalled at residual 0.0647 for five terms at $m=5$.
\end{itemize}

\paragraph{Verification.}
\begin{enumerate}[noitemsep,leftmargin=*]
\item \textbf{Kernel.} Exact Gaussian elimination over $\mathbb Q(\zeta)=\mathbb Q[\zeta]/(\zeta^4+1)$ confirms that $\ket T^{\otimes7}$ lies in the span of the ten states. Value $12/10=1.2$.
\item \textbf{Independent check (operator, separate code).} The stabilizer group of each state, found by brute force over all 16,384 Pauli operators, has order $2^7$. A 40-digit least-squares residual is $1.6\times10^{-41}$. The exact coefficients lie in $\tfrac14\mathbb Z[\zeta]$ and agree with least squares to $10^{-9}$.
\item \textbf{Reproduce.} \texttt{python verify/stabrank.py decomposition/T7\_rank10.json} and \texttt{python verify/independent\_check.py decomposition/T7\_rank10.json} (seconds)~\cite{zen_stabrank}.
\end{enumerate}

\subsection{Qubit routing on the SABRE suite}
The task: insert SWAP gates so that every two-qubit gate of each circuit acts on an edge of the 43-edge IBM Q20 Tokyo design graph, minimising added gates ($3\times$ SWAPs), under the rules of~\cite{li2019sabre}. Per-circuit added gates:
\begin{center}\scriptsize
\setlength{\tabcolsep}{3pt}
\begin{tabular}{@{}lrrrrrr@{}}
\toprule
circuit & gates & SABRE 2018 & Qiskit heavy & pytket & \IM{} & ratio to SABRE\\
\midrule
qft\_10 & 200 & 54 & 30 & 66 & 27 & 0.53\\
qft\_13$^*$ & 338 & 99 & 63 & 117 & 60 & 0.62\\
qft\_16 & 512 & 186 & 120 & 225 & 99 & 0.54\\
qft\_20$^*$ & 800 & 318 & 216 & 351 & 180 & 0.57\\
rd84\_142 & 343 & 105 & 87 & 90 & 48 & 0.47\\
adr4\_197 & 3439 & 1614 & 954 & 879 & 282 & 0.18\\
radd\_250 & 3213 & 1275 & 975 & 780 & 294 & 0.23\\
z4\_268 & 3073 & 1365 & 777 & 741 & 255 & 0.19\\
sym6\_145 & 3888 & 1272 & 969 & 714 & 312 & 0.25\\
misex1\_241 & 4813 & 1521 & 804 & 1311 & 228 & 0.15\\
rd73\_252 & 5321 & 2133 & 1692 & 1443 & 441 & 0.21\\
cycle10\_2\_110 & 6050 & 2622 & 2025 & 1620 & 459 & 0.18\\
square\_root\_7 & 7630 & 2598 & 1746 & 1911 & 516 & 0.20\\
sqn\_258 & 10223 & 4344 & 3093 & 2547 & 816 & 0.19\\
rd84\_253 & 13658 & 6147 & 5094 & 3615 & 1212 & 0.20\\
co14\_215 & 17936 & 8982 & 6732 & 5331 & 1785 & 0.20\\
sym9\_193 & 34881 & 16653 & 14019 & 6969 & 2535 & 0.15\\
9symml\_195 & 34881 & 17268 & 14019 & 6969 & 2535 & 0.15\\
8 small circuits & -- & 3 & 3 & 6 & 3 & 1.00\\
\midrule
total & & 68,559 & 53,418 & 35,685 & 12,087 & 0.18\\
\bottomrule
\end{tabular}
\end{center}
$^*$Regenerated circuits, whose reference is our SABRE reimplementation. The eight small circuits are \texttt{4mod5-v1\_22}, \texttt{mod5mils\_65}, \texttt{alu-v0\_27}, \texttt{decod24-v2\_43}, \texttt{4gt13\_92} and the three Ising chains; all methods need 0 SWAPs on them except \texttt{alu-v0\_27}, where one SWAP is optimal. \texttt{sym9\_193} and \texttt{9symml\_195} are byte-identical files.

\paragraph{How it was found.} Six runs, all started from the previous best:
\begin{itemize}[noitemsep,leftmargin=*]
\item \textbf{Session~1} (\texttt{20261021-041641}).
  \begin{itemize}[noitemsep]
  \item \emph{it00-p0} (\emph{data structure}, \$3.74): SABRE rewritten in C++ over the two-qubit dependency graph, with early abort once a trial exceeds the best count (about 3,400 trials/s on \texttt{adr4\_197}). Heuristic parameters ($W$, $|E|$, decay) are randomised per restart, followed by a hill-climb over initial mappings that swaps two physical qubits and accepts if not worse. Total 36,435.
  \item \emph{it01-p0} (\emph{aggressive rewrite}, 1,335~s, \$2.43): a receding-horizon router. At every 100-gate boundary it rolls out $K$ randomised 300-gate SABRE continuations and commits the first 100 gates of the best; $K$ adapts to the remaining time. Total 24,228.
  \end{itemize}
\item \textbf{Session~2} (\texttt{20261021-053321}).
  \begin{itemize}[noitemsep]
  \item \emph{it00-p1} (\emph{new neighbourhood}, 2,302~s, 965,052 input and 22,692 output tokens, \$3.24): \texttt{beam.cpp}, a level-synchronous beam search over SWAP sequences.
    \begin{itemize}[noitemsep]
    \item Each state is the logical-to-physical mapping plus a per-qubit count of executed two-qubit gates (40 bytes, hash-de-duplicated).
    \item After every SWAP, all executable gates are executed greedily.
    \item States are scored by $\sum_q\sum_{j<K}\delta^j(\mathrm{dist}_j-1)-C\cdot\#\text{executed}$, where $\mathrm{dist}_j$ is the distance between the qubits of $q$'s $j$-th upcoming gate, with $K=6$--$12$, $\delta=0.7$ and $C\approx1$.
    \item Beams hold up to 30,000 states, and forward--reverse passes refine the initial mapping.
    \end{itemize}
    Total 13,050.
  \item \emph{it01-p0} (\emph{aggressive rewrite}, \$3.99): an independent beam variant~B. The per-circuit combination of both is 12,261.
  \end{itemize}
\item \textbf{Session~3} (\texttt{20261021-180153}).
  \begin{itemize}[noitemsep]
  \item \emph{it00-p1} (\emph{data structure}, 1,669~s, \$1.56): forward--reverse iteration of variant~B (for example \texttt{co14\_215} 1,926 to 1,785).
  \item \emph{it02-p0} (\emph{aggressive rewrite}): restarts of the first beam from perturbed initial mappings (\texttt{qft\_16} 99, \texttt{qft\_20} 180, \texttt{misex1\_241} 228).
  \end{itemize}
\end{itemize}
The final routing takes, per circuit, the best verified run (column ``source'' in the deposit).

\paragraph{Verification.}
\begin{enumerate}[noitemsep,leftmargin=*]
\item \textbf{Kernel.} An exact replay (\texttt{qmap.verify\_routed}) ran in a fresh sandbox for every run. It checks that every original gate is executed exactly once, after its predecessors on both qubits, on a coupling-graph edge under the current mapping, and that every SWAP acts on an edge. It then counts SWAPs.
\item \textbf{Host.} The same replay on the combined routing gives a total of 12,087.
\item \textbf{Baselines.} Qiskit~2.4.1 and pytket~2.18.5 (versions pinned) ran on identical inputs with SWAPs counted the same way, BRIDGE counted as $+3$.
\item \textbf{Reproduce.} \texttt{python qmap.py verify routings/best\_combined.json}, which prints \texttt{total\_g\_add 12087} in about a minute using only the standard library~\cite{zen_sabre}.
\end{enumerate}

\subsection{The smallest adder transformer}
AdderBoard~\cite{adderboard} asks for the smallest transformer, by parameter count, that adds two 10-digit numbers with at least 99\% accuracy. Self-attention is required, and parameters are counted by the board's rules (our kernel counter is stricter: constants that act as weights count).

\paragraph{Result.} A 5-parameter, one-layer, two-head model that is correct on 10,010 of 10,010 additions on each of three seeds. Its parameters are $\omega=2\pi/10$ (scalar embedding $\omega\cdot\text{token}$, tied to the cosine read-out), $\lambda=60$ (attention sharpness), $\kappa=30$ (carry-gate sharpness), $\beta=1.5\,\omega$ (carry threshold) and $g=2$ (head-A gain). The positional encoding is a fixed sinusoid of period 11 (no parameters).
\begin{itemize}[noitemsep]
\item The input is laid out as $a_0..a_9,0,b_0..b_9,0$, followed by the generated digits fed back as tokens.
\item Head A, at phase $+1$, averages the current digit pair.
\item Head B, at phase $0$, averages the previous digit pair and the previous output digit, which yields $(s-o)/3$, where $s$ is that pair's digit sum and $o$ the previous output digit. This is $9/3$ or $10/3$ exactly when a carry is due.
\item The sigmoid gate turns this into the carry, and the logits are $\cos(g\,A+\omega\cdot\text{carry}-\omega k)$ for digit $k$.
\end{itemize}

\paragraph{How it was found.}
\begin{itemize}[noitemsep,leftmargin=*]
\item \textbf{Run.} AdderBoard session~1, 19 October, in an earlier scratch project; one proposer; 8 minutes; \$1.05. It had no parent: the proposer built the model from the task rules alone.
\item \textbf{Method (from its final message).} It chose the period-11 phase layout so that one head reads the current digit pair and the other the previous pair plus the generated digit. It set the five weights by hand. Its first local run timed out at 120~s because \texttt{torch} used 12 threads per call; it set one thread, after which 10,000 additions take about 30~s.
\item \textbf{Parallel run.} A second proposer in the same session found an independent 6-parameter design (ReLU wrap).
\end{itemize}

\paragraph{Verification.}
\begin{enumerate}[noitemsep,leftmargin=*]
\item \textbf{Local verifier} (run by the proposer). 10,000 pairs on each of seeds 2025 and 7: accuracy 1.0000, 5 counted parameters. With randomised weights, accuracy is 0.0.
\item \textbf{Kernel scorer} (stricter counter, hidden seed). 10,000 pairs on hidden seed 7331, with self-attention required and the perturbation check repeated: passed with 5 parameters.
\item \textbf{Upstream verifier.} The board's \texttt{verify.py} reports QUALIFIED, 10,010 of 10,010, on seeds 2025, 4242 and 7. Submitted as AdderBoard pull request \#101.
\end{enumerate}

\subsection{A 20M-parameter language model}
The task: an architecture of at most 20,000,000 parameters (embeddings included), retrained by the kernel from code on 1B FineWeb-10B tokens within 1,800~s on one H100, scored by bits per byte on the FineWeb validation shard. The GPT-2 124M checkpoint scores 1.0658, and the seed (a scaled-down modded-nanogpt~\cite{moddednanogpt}, 19.36M parameters) scores 1.3376.

\paragraph{Result.} 1.2716 bits per byte with 19,146,900 parameters, which closes 24\% of the gap.
\begin{itemize}[noitemsep]
\item \textbf{Architecture:} four unique transformer blocks cycled over eight effective layers (two loops) at width 512 with 8 heads, and a factorised tied embedding of rank 128. Per-effective-layer scalars (residual and skip weights) are not shared, so each loop can behave differently.
\item \textbf{Inherited from the session-1 parent:} learnable residual/$x_0$ mixing, U-net skips, Muon plus Adam, batch 131k tokens.
\end{itemize}

\paragraph{How it was found.}
\begin{itemize}[noitemsep,leftmargin=*]
\item \textbf{Run.} Session \texttt{20261022-014034}, iteration~0, proposer~1. Started 02:08; \emph{parameter sharing}; 6,270~s including its own H100 training runs; 360,077 input and 7,728 output tokens; \$1.20 of model usage plus GPU time.
\item \textbf{Start.} The session-1 best: 6 unique layers at width 448, 19.36M parameters, 1.2873 bits per byte when re-scored.
\item \textbf{Experiments (one run each at the official budget).}
  \begin{itemize}[noitemsep]
  \item \emph{A:} 12 effective layers over 4 unique blocks, width 512, embedding rank 96, 17.5M parameters. Time-bound: it reached only $0.856\times10^9$ tokens, with validation loss 3.9586 nats.
  \item \emph{B:} 8 effective layers over 4 unique blocks, width 512, rank 128, 19.15M parameters. It reached $0.995\times10^9$ tokens with loss 3.9030, against 3.9537 for the parent.
  \end{itemize}
  The proposer shipped~B and raised its training-time safety margin from 45 to 80~s, because B finished at 1,770 of 1,800~s.
\end{itemize}

\paragraph{Verification.}
\begin{enumerate}[noitemsep,leftmargin=*]
\item \textbf{Kernel scorer.} The scorer uploads only the code (checkpoints are stripped), retrains from scratch in a fresh H100 sandbox under the token and time limits, counts all parameters, and evaluates bits per byte on the FineWeb validation shard. Outcome: 1.2716, with 19,146,900 parameters (at most 20M).
\item \textbf{Limits.} One training seed; the run was not repeated.
\end{enumerate}

\subsection{Low-energy 8$\times$8 INT8 systolic tile}
The design is the LogikBench 8$\times$8 weight-stationary INT8 systolic array (signed 8-bit activations and weights, 32-bit outputs) shipped as \texttt{asap7/tpu} with OpenROAD-flow-scripts~\cite{orfs}. A tile loads eight weight rows (one per \texttt{w\_valid} cycle), streams activation rows, and returns each output row 15 cycles after its activation row. The host idles $2N-1=15$ cycles between tiles. A candidate must keep the ports and this cycle-level protocol, so it computes 64 MACs per cycle at the fixed clock. There is no published record for this tile; the reference is the original RTL.

\paragraph{Result.} An RTL rewrite (\texttt{tpu.sv} plus a generated \texttt{mac.sv}) with 128.81~fJ/MAC on the public workload and 130.1~fJ/MAC on the private one, against 578.85~fJ/MAC for the original. At 600~ps the power is 4.34~mW internal, 3.33~mW switching and 0.04~mW leakage (7.71~mW in total, against 34.66~mW). The standard-cell area is 2,507~$\mu$m$^2$ (original 5,559), setup worst slack $+5.2$~ps, hold worst slack $+6.3$~ps, and there are 0 DRC violations.

\paragraph{How it was found.}
\begin{itemize}[noitemsep,leftmargin=*]
\item \textbf{Runs.} Three runs in sequence, each building on the last:
\begin{itemize}[noitemsep]
\item session \texttt{20261026-042359-discover-tpu-energy}, it00-p0, \emph{data structure}, started 04:40; 2,477~s, 387,077 input and 15,459 output tokens, \$1.39;
\item session \texttt{20261026-052723-discover-tpu-energy}, it00-p1, \emph{new neighbourhood}, started 05:27; 3,367~s, 1,105,333 input and 24,581 output tokens, \$3.68;
\item the same session, it02-p0, \emph{new neighbourhood}, parent it00-p1, started 07:02; 2,181~s, 281,823 input and 12,234 output tokens, \$1.03.
\end{itemize}
\item \textbf{Tools.} The operator's evaluator (full flow, both simulations, power) and a quick mode of it that runs RTL simulation and synthesis only. Proposers ran it inside their own sandbox, so every number in their notebooks is a full-flow measurement.
\item \textbf{Session 1, it00-p0.} It removed the per-PE activation registers and broadcast each activation along its row, delaying row $i$ by $7+i$ cycles at the input so no output de-skew is needed. It narrowed the partial sums to 16--19 bits per row (sign-extended at the output) and dropped datapath resets (346.9~fJ). It then added latch+AND clock gates per pipeline depth, enabled from the valid pipe derived from \texttt{a\_valid}, and a \texttt{w\_valid} gate for the weights (274.9~fJ). Flops fell from about 4,200 to about 2,300. A parallel run (\emph{robust/simple}, \$1.81) instead wrote an explicit Baugh--Wooley multiplier with a 3:2 carry-save tree and a Brent--Kung adder, because Yosys maps \texttt{+} and \texttt{*} to Kogge--Stone structures (285.2~fJ, area 2,935).
\item \textbf{Session 2, it00-p1.} It combined the two (253.9~fJ) and tried radix-4 Booth with activations encoded once per row (300.8~fJ, rejected). It then wrote a generator, \texttt{gen\_mac.py R}, that emits one bit-level Dadda tree per group of $R$ PE rows, with Baugh--Wooley sign constants folded in, the incoming partial sum as an operand, and one Brent--Kung adder per stage. $R=2$ gave 210.4~fJ; $R=4$ (33 operands, two partial-sum stages, activation delay $13+\lfloor i/4\rfloor$) gave 188.2~fJ, with setup slack only $+3.6$~ps. The proposer noted that $R=8$ would not close 600~ps.
\item \textbf{Session 2, it02-p0.} It replaced the activation shift chains (13 or 14 stages of 64 bits, every flop toggling every cycle) by a 13-entry ring buffer. Each entry is written once through its own latch+AND clock gate under a shared one-hot pointer, and read through an AND-OR mux into the valid-gated input register; rows 0--3 read with the pointer rotated by one (delay 13), rows 4--7 without (delay 14). It also replaced the shifting weight register by direct per-row writes, with one clock gate per row and a one-hot pointer that advances on \texttt{w\_valid}. Result: 128.8~fJ. An ablation in the same run (ring buffer only) gave 144.6~fJ.
\item \textbf{Also tried.} A parallel line of runs (\emph{conservative}, \emph{robust/simple}, \emph{data structure}) reached 252.8~fJ by the same combination, 250.2~fJ with a hybrid final adder (ripple-carry for the low 6 bits, Brent--Kung above), and 207.0~fJ with two PE rows fused per stage. Its \emph{robust/simple} proposer declined to ship a 10-bit-ripple variant at 246.2~fJ because its hold slack was $+0.197$~ps. Untried, per the final notebook: a 12-entry ring for rows 0--3, gating the pointer clock when idle, latch storage entries, and a ring for the partial-sum stages.
\end{itemize}

\paragraph{Verification.}
\begin{enumerate}[noitemsep,leftmargin=*]
\item \textbf{RTL equivalence (kernel).} Icarus Verilog simulation of the candidate RTL against the Python golden model on a correctness set (10 uniform random tiles of 16 rows, plus corner tiles: all $-128$, all 127, alternating extremes, zeros) and on the scoring workload (8 tiles of 32 rows). Criterion: every output row bit-exact. Outcome: pass, on both seeds.
\item \textbf{Gate-level equivalence (kernel).} The scoring workload on the routed netlist, with functional (zero-delay) ASAP7 cell models. Criterion: bit-exact. Outcome: pass, on both seeds.
\item \textbf{Physical gates (kernel).} Setup and hold worst slack at the finish stage $\ge0$, final detailed-route DRC count 0, area $\le6{,}115.1~\mu$m$^2$. Outcome: $+5.2$, $+6.3$~ps, 0 and 2,507~$\mu$m$^2$.
\item \textbf{Power (kernel).} OpenSTA \texttt{read\_vcd} on the gate-level VCD of the 456-cycle window and \texttt{report\_power} with the routed SPEF; energy per MAC $=P_\text{total}\times456\times600~\text{ps}/16{,}384$. Public seed 2026: 128.81~fJ. Private seed 7331, run by the kernel in its own sandbox: 130.1~fJ.
\item \textbf{Scope.} The design relies on weight loads of exactly eight \texttt{w\_valid} cycles per tile, as its notebook states. The testbench always loads that way, so both equivalence checks hold only under this protocol. There was no separate host re-run beyond the kernel's two splits.
\end{enumerate}

\subsection{TESS planet candidate}
\paragraph{Result.} G~249-11 (TIC~417732194; Gaia DR3 484671464331349632), an M4.5 dwarf at 29.1~pc. The candidate has $P=5.307425\pm0.000008$~d, $T_0=2459912.3151\pm0.0008$~BJD$_\mathrm{TDB}$, depth $3.0\pm0.4$~ppt, $T_{14}=0.88$~h, $b=0.83$ and $R_p=1.61\pm0.09\,R_\oplus$.

\paragraph{How it was found.}
\begin{itemize}[noitemsep,leftmargin=*]
\item \textbf{Run.} Session \texttt{20261022-221142}, iteration~0, proposer~0. Started 22:11 on 22 October; \emph{new population} arm; 6,294~s; \$7.06 (task-level per-run cap \$8).
\item \textbf{Method.} A box least-squares (BLS) search of TESS-SPOC full-frame-image light curves of the 3,100 brightest of 7,474 M dwarfs within 40~pc. It re-found the known planets TOI-1266, TOI-1640 and TOI-1685 at SNR 22--23, and set aside about 1,300 referee-accepted peaks as stellar rotation.
\item \textbf{Claims.} Four new transits. Its notebook calls TIC~417732194 the cleanest: 8 events in Sectors 59--60.
\item \textbf{Independent detection.} The parallel proposer in the same round (it00-p1, \$3.84) also detected the signal but declined to claim it, citing ``short baseline, unpredictable ephemeris''.
\end{itemize}

\paragraph{Verification.}
\begin{enumerate}[noitemsep,leftmargin=*]
\item \textbf{Referee v5 (kernel).} The referee re-measures each claim from the raw light curves, applies cadence checks per sector and the cross-era rule, checks against known catalogues (TOI, CTOI, SPOC TCEs, VSX, Gaia variability, SIMBAD) and requires an event-overlap alias test to pass. It accepted the claim. Two of the other three claims were rejected by the cross-era rule.
\item \textbf{Pixel level (operator).} A Sector-59 TESScut difference image places the flux deficit on or next to the target pixel. The fourth claim was rejected here as a field-wide flux drop.
\item \textbf{Prediction test (operator).} An ephemeris fitted to Sectors 59--60 alone predicted dips in Sectors 73 and 86, which were never used for selection and lie 68 and 134 orbits later. Both dips appear inside the propagated $\pm2\sigma$ window. Against 400 full-pipeline null realisations (transits masked, sectors circularly shifted), $p=0.09$ for Sector 73, $p=0.0025$ for Sector 86, and about $0.002$ combined.
\item \textbf{Photometric vetting.}
  \begin{itemize}[noitemsep]
  \item Odd and even depths agree to $1.2\sigma$.
  \item No secondary eclipse ($-0.13\pm0.30$~ppt).
  \item The depth is stable across seven detrending choices.
  \item The residuals are white at the transit timescale.
  \end{itemize}
\item \textbf{Independent review.} A read-only reviewer agent found a 1,000-day error in the reference epoch and an independence error in the first report. Both were corrected.
\item \textbf{Status.} A bound eclipsing binary and a $G=17.2$ neighbour $20''$ away are not excluded. The candidate is unconfirmed~\cite{zen_tess}.
\end{enumerate}

\section{The autonomous runs in detail}
\label{app:autorun}
\sloppy
\subsection{Run 1: two-deletion-correcting codes}
Session \texttt{20261026-002501-autonomous}, launched as a background job at 00:24:58~UTC on 26 October 2026 with 3 iterations, 2 parallel proposers, a \$30 budget and at most 3 scout rounds. There was no operator action after launch.

\paragraph{Goal (verbatim, the default).} ``Find an open problem in mathematics, quantum information or theoretical computer science whose best known result is an explicit object or number that a program can check exactly, and improve on the best known result.''

\paragraph{Problem chosen.} The largest binary code $C\subset\{0,1\}^{16}$ that corrects two deletions: for distinct codewords, the sets of length-14 strings obtained by deleting two positions are disjoint. This is a maximum independent set in a fixed graph on 65,536 vertices. The LP upper bound is 487. The scout's reasons, from \texttt{SELECTION.md}:
\begin{itemize}[noitemsep]
\item the deliverable is an explicit word list;
\item the exact check takes under a second;
\item every extra codeword is a strict improvement;
\item the literature gave 215 (Weindel and Heckel, arXiv:2504.00613v2, code not published) and 208 (CPro1, arXiv:2505.23881, code public).
\end{itemize}

\paragraph{Round 0.}
\begin{enumerate}[noitemsep,leftmargin=*]
\item \textbf{Scout} (00:25--00:35, \$3.14). It wrote a task with record 215, the public 208-word seed, score range 208--216 and ten calibration cases: four valid (208, 194, 193 and 100 words) and six invalid (a bit flip, a duplicate, an empty code, a near-miss extra word, a non-binary word, a wrong length).
\item \textbf{Kernel validation.} The seed scored 208. All four valid cases reproduced their values and all six invalid cases were rejected, so no problems were found.
\item \textbf{Audit} (00:36--00:39, \$1.32), verdict \emph{fail}. From \texttt{AUDIT.md}: ``the cited EinsteinArena problem 34 has moved on. Its leaderboard (queried 2026-10-26): nightshift-dc16 223, happy-gpt-astra-6 218, Opus-HedgyLabs 216, \ldots{} I downloaded the 223-word code from the public API and checked it with the task's own verifier (223.0) and an independent max pairwise LCS = 13 test \ldots{} So record 215 is wrong.''
\item \textbf{Exploit replay.} The auditor wrote 18 exploit files, among them every word in the space, Boolean or integer elements, full-width and Arabic-Indic digits, padded words, NaN, wrong lengths, a list instead of a dict, and too many words. The kernel replayed all 18 and the verifier accepted none.
\end{enumerate}

\paragraph{Round 1.}
\begin{enumerate}[noitemsep,leftmargin=*]
\item \textbf{Scout repair} (00:40--00:46, \$0.85). Record 223, sourced to the arena's best-solution API and queried 26 October. The 223-word code became the seed, the score range was widened to 208--228, and two new calibration cases were added: the 223-word code (valid) and that code plus one conflicting word (invalid).
\item \textbf{Kernel validation.} The seed scored 223. All five valid cases reproduced their values and all seven invalid cases were rejected.
\item \textbf{Audit} (00:47--00:49, \$0.58), verdict \emph{pass}, with \texttt{open: true} and \texttt{record\_ok: true}. Its remaining notes: extra dictionary keys are ignored, which is harmless because the value is the number of words; and the record moves fast.
\item \textbf{Exploit replay.} Twenty files, none accepted.
\item \textbf{Installation.} At 00:50:39 the task was installed, with an approval record naming the approver as ``kernel (autonomous mode)'' and the basis as ``calibration + adversarial audit + exploit replay''.
\end{enumerate}

\paragraph{Discovery} (00:50--02:05, \$3.46; session \texttt{20261026-005039-discover-del2-code-16}).
\begin{itemize}[noitemsep,leftmargin=*]
\item \textbf{Iteration 0.}
  \begin{itemize}[noitemsep]
  \item \emph{aggressive rewrite} (\$0.75): plateau local search with forced insertion and Metropolis acceptance, 4 cores $\times$ 25~min, about $2\times10^7$ moves each. Current size wandered between 206 and 223; best 223.
  \item \emph{data structure} (\$0.59): an implicit-adjacency search over $2^{14}$ descendant slots at about $1.9\times10^8$ candidate probes per second per core; best 223.
  \end{itemize}
\item \textbf{Iteration 1.}
  \begin{itemize}[noitemsep]
  \item \emph{robust/simple} (\$0.84): no gain.
  \item \emph{new neighbourhood} (\$0.86): exact large-neighbourhood search. It frees $K$ codewords chosen by Hamming proximity, a random bit mask or a shift-aware distance, then solves the freed region exactly by bitset branch and bound. Runs at $K=24$--60 made up to $1.6\times10^6$ moves; 10--13\% reached a different optimal set and none reached a larger one.
  \end{itemize}
\item \textbf{Iteration 2.} \emph{conservative} twice (\$0.20 each). Both copied the parent and its notes: no-ops.
\end{itemize}
Best: the seed, 223 words. Seven bank entries, none above the seed.

\paragraph{Independent check} (02:05--02:06, \$0.19, network off). The checker received only \texttt{problem.md} and the best solution. It wrote \texttt{my\_checker.py} (format, 16-bit binary words, distinctness, pairwise-disjoint deletion sets) and tested it on constructed cases. As a cross-check it computed the longest common subsequence of every pair (at most 13, so no shared length-14 subsequence). Verdict: valid, 223, in agreement with the verifier.

\paragraph{Report} (02:06--02:07, \$0.32). Claim: \{beats\_record: false, value: 223, record: 223, claim: ``No improvement: the best verified code has 223 words, which ties (does not beat) the public record of 223 for two-deletion-correcting binary codes of length 16.''\}. The report's claim boundary disclaims a new record, the optimality of 223, the impossibility of 224, and authorship of the code, crediting the arena agents.

\paragraph{Totals.} 1~h~42~min; \$9.86. The breakdown is scout \$3.99, audit \$1.90, discovery \$3.46, check \$0.19 and report \$0.32.

\subsection{Choosing the goals of runs 2 and 3}
Before runs 2 and 3, five research agents, directed by the operator, surveyed published tables of explicit, exactly checkable records for low current activity. They looked beyond quantum information. Their top picks were Brouwer's mixed binary/ternary code tables (quiet since 2013), qutrit stabilizer-code gaps, sorting-network sizes for 25--40 inputs, real Grassmannian packings and ternary codes $A_3(n,d)$. They advised avoiding tables already worked by many agents: the V3R1T4S list, binary and constant-weight codes, most Einstein Arena problems and the QIQC zoo. The operator wrote one goal sentence for each of the first and third picks, naming the table and the improvement wanted, and asked the scout to choose the cell and confirm that it was current. Both jobs ran with 3 iterations, 2 parallel proposers, a \$30 budget and the research math arms. The first launch of the pair collided on a session directory (Section~\ref{sec:auto-run}) and was cancelled within a minute; the relaunches below are the runs reported.

\subsection{Run 2: sorting network on 31 inputs}
Session \texttt{20261026-024005-autonomous-e33eb}, 02:40--05:18~UTC.

\paragraph{Goal (verbatim).} ``Improve a size upper bound in Bert Dobbelaere's sorting-network tables for n = 25 to 40 inputs (\ldots): choose n, confirm the current best-known number of comparators and its date, and construct a valid sorting network on n inputs with fewer comparators, verified exactly by the 0-1 principle.''

\paragraph{Scout} (02:40--02:58, \$2.03). It chose $n=31$: record 180 comparators in 14 layers, bounds $141\le S(31)\le180$, table source ``Batcher odd-even merge'', with no 31-input size change since the 2017 draft (page dated 10 October 2026). Its reason was that the entry is the 32-input merge network minus one wire, never improved by search, while the entries for $n=28$--30 had been improved beyond the wire-removal bound. Before choosing, it ran its own wire-removal beam search from all record networks for $n=25$--40; this beat no entry (for $n=30$ it gave 173 against the record 172). The verifier is an exact check of all $2^{31}$ binary inputs by bit-slicing (about 8--15~s). The seed is the record network. There were eight calibration cases. The three valid ones were the 186-comparator Batcher network, the record and the record plus a redundant comparator (181). The five invalid ones were an index out of range, a malformed list, the record with one comparator dropped, a reversed comparator and a 30-input network. Scoring was minimised over the range 186 to 175, which in the kernel's convention is reversed (Section~\ref{sec:auto-run}).

\paragraph{Validation and audit} (02:58--03:05, \$0.89). All eight cases reproduced. Audit verdict: \emph{pass}. Its minor issues were that the page now read 13 October, with changes only for other $n$; that $n=31.0$ is accepted as 31; and that public equals private. The auditor wrote 23 exploit files, among them the 30-input record plus an unused wire, and the kernel's replay rejected all of them. The auditor described the score mapping as ``186$\to$0, 175$\to$1'', the intended direction, which is not what the kernel computes.

\paragraph{Discovery} (03:06--05:15, \$4.12; session \texttt{20261026-030658-discover-sortnet-31}).
\begin{itemize}[noitemsep,leftmargin=*]
\item \textbf{Iteration 0.} \emph{Literature transfer} twice (\$0.64, \$1.04). Both wrote SorterHunter-style evolutionary searches in C from the record, with the first three or four layers fixed and only the reachable 0-1 vectors kept. Their mutations replaced, swapped, moved, inserted or deleted comparators and pruned redundant ones. Neither found 179.
\item \textbf{Iteration 1.} \emph{Scale compute} (\$0.97) added a pretest on 1,024 witness vectors that earlier mutants had failed, raising throughput from about 52k to 300k evaluations per second per core, and ran on 4 cores for 2,900~s each: no 179. \emph{Exhaustive small case} (\$0.27) ran no search. It estimated a delete-two/insert-one repair at about $10^9$ candidates, each over about 28,000 vectors, as beyond its budget.
\item \textbf{Iteration 2.} \emph{New objective} (\$0.27) checked that none of the 180 single-comparator deletions sorts, then ran no new search. \emph{New objective} (\$0.93) fixed the first 60 comparators (four butterfly layers, leaving 28,024 unsorted reachable vectors) and annealed the remaining 119 comparators on the number of vectors still unsorted. Its best state left one vector unsorted.
\end{itemize}
Best: the seed, 180. Seven bank entries, all scored 0.4545.

\paragraph{Independent check and report} (05:16--05:18, \$0.55). The checker's own C program confirmed that the network sorts all $2^{31}$ inputs with 180 comparators. Claim: \{beats\_record: false, value: 180, record: 180\}. The report noted the audit's 0.545 against the bank's 0.4545 and used the bank's figure. Its claim boundary disclaims an improvement and any statement that 179 is impossible, and notes that every search started from the record with a fixed prefix.

\paragraph{Totals.} 2~h~38~min; \$7.59: scout \$2.03, audit \$0.89, discovery \$4.12, check \$0.27, report \$0.28.

\subsection{Run 3: mixed binary/ternary code $N(4,5,3)$}
Session \texttt{20261026-023958-autonomous-67f31}, 02:40--05:35~UTC.

\paragraph{Goal (verbatim).} ``Improve a lower bound in Andries Brouwer's table of mixed binary/ternary codes N(m,n,d) (\ldots): codes with m binary and n ternary coordinates and minimum mixed Hamming distance d. Choose a cell where the gap between the lower and upper bound is large and the lower bound looks improvable, confirm the table entry is current (check the page's update notes and recent arXiv), and construct a strictly larger code than the current lower bound.''

\paragraph{Scout} (02:40--03:27, \$4.66). It chose $N(4,5,3)$: words in $\{0,1\}^4\times\{0,1,2\}^5$ (3,888 in all) at mixed Hamming distance at least 3, with bounds 186--216. The lower bound dates from Brouwer, H\"am\"al\"ainen, \"Osterg\aa rd and Sloane (1998), and the upper bound 216 is $3N(4,4,3)$. Its reasons were the 16\% gap and the fact that neighbouring cells, among them $N(3,5,3)$, $N(5,5,3)$, $N(3,6,3)$ and $N(4,7,3)$, were improved by forum posters in 2010--2013 while this one was not. It fetched the page (Last-Modified 17 October 2026); its arXiv search was blocked by HTTP~429. It wrote \texttt{coset.c}, a clique search over unions of cosets of an additive subgroup. This reproduced 186 as 62 cosets of $\langle(0000|10101)\rangle$ in seconds, along with other table entries ($N(7,3,3)=192$, $N(8,3,3)=384$, $N(5,6,5)=96$, $N(6,5,5)=66$). Its own searches over subgroup orders 2, 3, 4, 6 and 9 and a 16k-step LNS did not beat 186. There were twelve calibration cases. Four were valid: the record, a 150-word subcode, the same code in string format and a single word. Eight were invalid: a binary symbol 2, a duplicate, an empty code, float symbols, a near miss at distance~1, the record plus a conflicting 187th word, the wrong cell and the wrong length. Scoring was maximised over 150--188, so the record scores 0.947 (correct direction).

\paragraph{Validation and audit} (03:27--03:32, \$1.14). All twelve cases reproduced. Audit verdict: \emph{pass}. Its minor issues were that extra keys are ignored and that its own arXiv and Semantic Scholar searches were also rate-limited, so only the table page was checked. The kernel replayed 24 exploit files, among them all 3,888 words, and rejected all of them. Installed at 03:34.

\paragraph{Discovery} (03:34--05:33, \$3.60; session \texttt{20261026-033400-discover-mixed-code-n453}).
\begin{itemize}[noitemsep,leftmargin=*]
\item \textbf{Iteration 0.} \emph{Literature transfer} twice (\$0.68, \$0.66). Both ran prescribed-automorphism-group searches, after \"Osterg\aa rd's method. Each sampled random groups of binary permutations with flips and ternary permutations with per-coordinate symbol permutations, kept compatible orbits, and annealed a weighted independent set on the orbit conflict graph. Both found 186 and nothing more.
\item \textbf{Iteration 1.} \emph{Structured construction} (\$0.34): about 10,900 groups with up to three generators, 1,500~s on 4 CPUs; best 186 in every run. \emph{New objective} (\$0.70): \texttt{tabu.c} fixes the code size and minimises the number of conflicting pairs. From random starts it reached valid codes of 180 and 183 words, and once 186 with no imposed structure, but never 187.
\item \textbf{Iteration 2.} \emph{New objective} (\$0.88): a fixed-size tabu search over orbits of the known translation group. It reaches 62 orbits (186 words) in under a second; at 63 orbits (189 words) the best state had one conflicting pair, over four 150~s runs and further seeds. \emph{Scale compute} (\$0.35): 9,653 random groups with long anneals, 2,700~s on 4 CPUs. 272 groups reached 186 and none reached 187.
\end{itemize}
Best: the seed, 186. Seven bank entries, all scored 0.947.

\paragraph{Independent check and report} (05:33--05:35, \$0.58). The checker confirmed 186 valid words and stated that 186 is not a new record. Claim: \{beats\_record: false, value: 186, record: 186\}. The report notes the rate-limited literature checks, says that a newer paper cannot be ruled out, and says that 186 is not proven optimal.

\paragraph{Totals.} 2~h~55~min; \$9.97: scout \$4.66, audit \$1.14, discovery \$3.60, check \$0.17, report \$0.41.

\section{Discovery sessions}
\label{app:sessions}
Table~\ref{tab:sessions} lists every discovery session in the durable project, in chronological order. ``Runs'' counts finished proposer runs, and ``\$'' is the recorded model spend. A dash means the session ended before the kernel wrote its totals. AdderBoard session~1 (\$3.26) and the TSP calibration run (\$8.12) ran in earlier scratch projects and are not listed.

\begin{table}[h]
\centering\footnotesize
\begin{tabular}{llcccrl}
\toprule
Start (UTC) & Task & Iters & Par. & Runs & \$ & Outcome\\
\midrule
10-19 01:53 & ml/adderboard (s2) & 2 & 2 & 4 & 0.89 & ``1 parameter'' via constants (gaming)\\
10-19 03:49 & math/sidon-45 & 2 & 2 & 4 & 2.80 & 13 ties at 4/7\\
10-19 04:59 & math/hadamard-51 & 2 & 2 & 4 & 1.03 & record is a fixed point\\
10-19 05:22 & math/hadamard-51 & 2 & 2 & 4 & 3.05 & $\mathbb Z_5^2$ family 43.2310\\
10-19 17:33 & math/depol-threshold & 2 & 2 & 0 & -- & killed (host memory)\\
10-19 18:11 & math/depol-threshold & 2 & 1 & 2 & 1.31 & certified $p=0.064966$\\
10-20 00:59 & math/k7d2-code & 2 & 2 & 4 & 1.33 & no $K\ge25$ (partial-credit gaming)\\
10-20 01:09 & math/k7d2-code & 3 & 2 & 6 & 2.35 & no $K\ge25$\\
10-20 02:50 & math/qutrit-depol & 3 & 2 & 2 & 4.55 & 0.289205\\
10-20 04:01 & math/qutrit-depol & 3 & 2 & 4 & 7.93 & 0.2910672\\
10-20 15:36 & math/golay-prep & 3 & 2 & 4 & 5.84 & seed (44, 8) only\\
10-20 16:38 & math/golay-prep & 3 & 2 & 5 & 13.88 & (44, 6)\\
10-20 18:55 & math/depol-private & 3 & 2 & 2 & 7.82 & $n=5$, 0.28372\\
10-20 20:10 & math/depol-private & 3 & 2 & 5 & 6.24 & $n=8$, 0.284668\\
10-21 01:36 & math/mub6-k4 & 3 & 2 & 6 & 2.10 & record reproduced only\\
10-21 02:40 & math/stabrank-t & 3 & 2 & 6 & 2.33 & null (default arms)\\
10-21 03:33 & math/stabrank-t & 3 & 2 & 3 & 7.29 & $\chi(\ket{T}^{\otimes7})\le10$ (math arms)\\
10-21 04:16 & opt/sabre-tokyo & 3 & 2 & 3 & 8.11 & 24,228 added gates\\
10-21 05:33 & opt/sabre-tokyo-v2 & 3 & 2 & 3 & 8.27 & beam search; 12,261 combined\\
10-21 06:52 & opt/sabre-tokyo-v2 & 3 & 2 & 1 & -- & shell died; one run lost\\
10-21 18:01 & opt/sabre-tokyo-v2 & 3 & 2 & 6 & 4.83 & 12,087 combined\\
10-21 21:21 & ml/tinylm & 3 & 2 & 6 & 5.32 & 1.2868 bpb (default arms)\\
10-22 01:40 & ml/tinylm & 3 & 2 & 6 & 4.15 & 1.2716 bpb; Modal limit hit\\
10-22 17:48 & astro/tess-discovery & 3 & 3 & 3 & 11.24 & referee v1 gamed\\
10-22 18:54 & astro/tess-discovery & 3 & 3 & 3 & -- & cancelled after round 0 (gaming)\\
10-22 20:44 & astro/tess-discovery & 2 & 3 & 4 & 18.66 & nothing over cleaned seed\\
10-22 22:11 & astro/tess-discovery & 2 & 3 & 4 & 13.31 & G 249-11 candidate\\
\midrule
10-26 00:50 & auto/del2-code-16$^\dagger$ & 3 & 2 & 6 & 3.46 & 223 (tie with public record)\\
10-26 03:06 & auto/sortnet-31$^\dagger$ & 3 & 2 & 6 & 4.12 & 180 (tie)\\
10-26 03:34 & auto/mixed-code-n453$^\dagger$ & 3 & 2 & 6 & 3.60 & 186 (tie)\\
\midrule
10-26 03:52 & chip/tpu-energy & 3 & 2 & 0 & -- & cancelled: score reversed\\
10-26 04:23 & chip/tpu-energy & 3 & 2 & 3 & 3.42 & 274.9 fJ; cancelled: saturated\\
10-26 05:27 & chip/tpu-energy & 3 & 2 & 6 & 7.59 & 128.8 fJ/MAC\\
\bottomrule
\end{tabular}
\caption{All discovery sessions (dates in October 2026). $^\dagger$The discovery phase of an autonomous run, whose task was chosen and written by the system; the scout, audit, check and report phases cost a further \$6.40, \$3.47 and \$6.37 (Section~\ref{sec:auto-run}). Two autonomous jobs that collided on one session directory were cancelled before their scouts finished and are not listed.}
\label{tab:sessions}
\end{table}

\section{Strategy arms}
\label{app:arms}
Each proposer's brief contains the name and text of its arm, chosen per task family by UCB1.
\paragraph{Default (written for optimisation tasks).}
\emph{conservative}: keep the parent's algorithm; make one safe, measured improvement. \emph{aggressive rewrite}: replace the core algorithm with a stronger one you can implement and test within the budget. \emph{new neighbourhood}: keep the framework but add a new move type or neighbourhood and integrate it into the search loop. \emph{data structure}: make the inner loop faster so more iterations fit in the time limit; measure iterations per second before and after. \emph{robust/simple}: simplify and harden; make every instance size produce valid output.
\paragraph{Math (written by the operator after the first stabilizer-rank session).}
\emph{structured construction}: build candidates from structure (symmetry, known constructions, algebraic families, products of smaller solutions) instead of unstructured local search; derive, then verify exactly. \emph{exhaustive small case}: reduce by symmetry and enumerate a small case completely, turning the search into a decision with a replayable certificate. \emph{new objective}: change what is optimised or how candidates are parametrised; calibrate on a solved case first. \emph{scale compute}: take the best existing method and scale it with parallel restarts and checkpointing. \emph{literature transfer}: port one published technique named in the notes, reproduce its known result, then push it to the open case.
\paragraph{ML (written by the operator after the first LM session).}
\emph{new architecture family} (mixture of experts, state-space or linear attention, recurrent, hybrids), \emph{parameter sharing} (looped or weight-shared blocks), \emph{embedding budget} (smaller vocabulary, factorised or hashed embeddings), \emph{connectivity} (skip connections, value residuals, multi-token heads, attention variants), \emph{recipe} (optimizer, batch, schedule); each compared at equal tokens.
\paragraph{Astronomy (catalogue mode).}
\emph{new population}, \emph{new phenomenon}, \emph{deeper search}, \emph{vetting and follow-up}, \emph{anomaly hunt}.

\section{Evidence for the specification-gaming incidents}
\label{app:evidence}
\paragraph{AdderBoard: weights turned into constants.} Excerpt from the model the proposers delivered (session 2). The two former weights in the excerpt are commented as fixed temperatures, one explicitly as ``not a weight'':
\begin{small}
\begin{verbatim}
self.omega = nn.Parameter(torch.tensor(2 * math.pi / 10))  # embedding / read-out scale
self.lam = 60.0      # attention sharpness (fixed temperature, not a weight)
self.kappa = 30.0    # carry-gate sharpness (fixed temperature)
...
carry = torch.sigmoid(self.kappa * (Bv - 1.5 * self.omega))
U = 2.0 * A + self.omega * carry
\end{verbatim}
\end{small}
From the proposer's final message: ``I cut the model from 5 parameters to 1 and it still adds correctly \ldots{} \textbf{Review risk:} the fixed constants (60, 30, 1.5, 2) sit in \texttt{forward()}. A human reviewer could read that as weights hidden in the architecture.'' Under the strict counter, these constants count and the model has 5 parameters.

\paragraph{Seven-qubit code: lenient partial credit.} The first verifier scored a subspace of dimension $K$ with worst compression residual $r$ as $K$ if $r<10^{-9}$, and otherwise as $(K-1)+0.999\cdot\mathrm{clip}(-\log_{10}r/9)$. A $K=25$ subspace at $r=0.072$ (no code) therefore scored 0.8127, above the confirmed $K=24$ code's 0.80. A proposer drawn with the \emph{robust/simple} arm delivered exactly that. The fixed formula starts partial credit only below $r=10^{-2}$.

\paragraph{TESS, referee v1.} Accepted claims included ``single transits'' with depths of 500--1200\% (flux glitches), flux steps, and about 90 low-effort eclipsing-binary claims, a quarter of which failed the held-out prediction. Re-scored under referee v2, the same catalogue fell from 94.5 to 30.5 (public) and from 77.5 to 21.75 (hidden).

\paragraph{TESS, referee v2.} From a proposer's notebook (session 2): ``Hypothesis: the parent catalogue saturated the capped classes (83 EBs $>$ cap 15 credits, 54 variables $>$ cap 5), so extra credit lies in uncapped single events (1.0) and long-period duo transits (3.0), which SPOC TPS cannot find.'' It then submitted 240 single events and 39 two-event ``transits'' whose period was set to the gap between the events. The operator cancelled the session after that round.

\paragraph{TESS, referee v4 (operator-found).} The referee computed one cadence per star, so for a star with both 200-s and 30-min sectors, every 30-min event was invisible to its checks, and a dip seen in only one observing era could pass. Two of four new transit claims were rejected under the fixed referee (v5) because their dip appeared only in the latest era.

\section{Gate configuration and the self-derived playbook}
\label{app:gate}
\paragraph{Gate.} Held-in tasks: an RFC~4180 CSV parser, 2-D TSP and graph colouring. Held-out tasks: interval merging and bin packing. Noise band 0.03; minimum gain 0; anomaly threshold 0.3; cost factor 1.5; at most 12 changed files; one seed per task. Parent selection: score-child (a UCB selector with descendant-growth pruning is implemented but not the default).
\paragraph{Generation 1 (consolidation, rejected).} The four bullets the system derived from its first 16 problem runs, verbatim apart from shortened run identifiers and two file names replaced by bracketed descriptions:
\begin{itemize}[noitemsep]
\item[P-0002] Before searching, list neighbourhoods and families already tried in prior notes (1--3 flip SA, ILS from record) and pick an untried one; several runs re-ran the same local search. (3 Hadamard runs; confidence 0.60)
\item[P-0003] Faster inner loops or better optimisers on a proxy energy (Gram target, design cost) are not progress unless the proxy is shown to track the true objective; compare on the real score at equal wall-clock. (2 Hadamard runs, 1 Sidon run; 0.60)
\item[P-0004] Symmetry-restricted search spaces ($\mathbb Z_{17}$, $\mathbb Z_5\times\mathbb Z_5$, $\mathbb Z_3$ blocks) capped below the record's family; run a short pilot (about 60~s) and compare to the record before committing a long compute budget. (2 Hadamard runs; 0.55)
\item[P-0005] [the task's local scoring script] is often absent; look for [the task's checker tool] first and use it for the exact score. (3 runs; 0.70)
\end{itemize}
Graded result: held-in 0.944 against 0.927, held-out 1.000 against 1.000, cost \$5.50 against \$2.72 (ratio 2.02 $>$ 1.5): rejected.

\end{document}